\pdfoutput=1  
\documentclass[sigconf, screen]{acmart}

\copyrightyear{2026}
\acmYear{2026}
\setcopyright{cc}
\setcctype{by}
\acmConference[MM '26]{Proceedings of the 34th ACM International Conference on Multimedia}{November 10--14, 2026}{Rio de Janeiro, Brazil}
\acmBooktitle{Proceedings of the 34th ACM International Conference on Multimedia (MM '26), November 10--14, 2026, Rio de Janeiro, Brazil}
\acmDOI{10.1145/3767308.3836424}
\acmISBN{979-8-4007-2213-4/2026/11}

\usepackage{algorithm}
\usepackage{algpseudocode}
\usepackage{multirow}
\usepackage{booktabs}
\usepackage[dvipsnames]{xcolor}

\newcommand{\cmark}[1]{\textcolor{ForestGreen}{#1}}
\newcommand{\xmark}[1]{\textcolor{red}{#1}}

\usepackage{enumitem}
\setlist{nosep, leftmargin=*, topsep=2pt, parsep=1pt}

\begin{document}

\title{\textbf{Latent Denoising Improves Visual Alignment in Large Multimodal Models}}

\author{Dhruv Parikh}
\authornote{Equal contribution.}
\affiliation{%
  \institution{University of Southern California}
  \city{Los Angeles}
  \country{USA}
}
\email{dhruvash@usc.edu}

\author{Jacob Fein-Ashley}
\authornotemark[1]
\affiliation{%
  \institution{University of Southern California}
  \city{Los Angeles}
  \country{USA}
}
\email{feinashl@usc.edu}

\author{Rajgopal Kannan}
\affiliation{%
  \institution{DEVCOM ARL Army Research Office}
  \city{Los Angeles}
  \country{USA}
}
\email{rajgopal.kannan.civ@army.mil}

\author{Viktor Prasanna}
\affiliation{%
  \institution{University of Southern California}
  \city{Los Angeles}
  \country{USA}
}
\email{prasanna@usc.edu}

\begin{abstract}
Large Multimodal Models (LMMs) such as LLaVA are typically trained with an
autoregressive language modeling objective, providing only indirect supervision
to visual tokens. This often yields weak internal visual representations and
brittle behavior under distribution shift. Inspired by recent progress on
latent denoising for learning high-quality visual tokenizers, we show that the
same principle provides an effective form of visual supervision for improving
internal visual feature alignment and multimodal understanding in LMMs. We
propose a latent denoising framework that corrupts projected visual tokens
using a saliency-aware mixture of masking and Gaussian noising. The LMM is
trained to denoise these corrupted tokens by recovering clean teacher patch
features from hidden states at a selected intermediate LLM layer using a
decoder. To prevent representation collapse, our framework also preserves the
teacher's intra-image similarity structure and applies intra-image contrastive
patch distillation. During inference, corruption and auxiliary heads are
disabled, introducing no additional inference-time overhead. Across a broad
suite of standard multimodal benchmarks, our method consistently improves
visual understanding and reasoning over strong baselines, and yields clear
gains on compositional robustness benchmarks (e.g., NaturalBench). Moreover,
under ImageNet-C-style non-adversarial common corruptions applied to benchmark
images, our method maintains higher accuracy and exhibits reduced degradation
at both moderate and severe corruption levels. Our code is available at
\url{https://github.com/dhruvashp/latent-denoising-for-lmms}.
\end{abstract}



\maketitle

\section{Introduction}
\label{sec:intro}

\looseness=-1 Large Language Models (LLMs), spanning proprietary systems such as GPT~5.4, Gemini~3.1~Pro, and Claude~4.6~Opus \cite{gpt-5.4, gemini-3.1-pro, claude-opus-4.6}, as well as open-source models such as Llama~4, DeepSeek-V2, and Qwen3-Next \cite{llama-4, deepseek-v2, qwen-3-next}, have rapidly become the backbone of modern AI systems and applications. Building on this progress, \emph{Large Multimodal Models} (LMMs) have emerged as a dominant paradigm for jointly reasoning over language and visual content. A large class of open-source LMMs follows a simple yet powerful recipe: pairing a pretrained LLM with a strong vision encoder through a lightweight projector and performing multimodal instruction tuning, as exemplified by LLaVA \cite{llava, llava-next, llava-improved} and its successors such as Qwen-VL and InternVL \cite{internvl, internvl-1.5, internvl-2.5, qwen-vl, qwen2-vl, qwen2.5-vl, qwen3-vl}. In parallel, another line of work explores models trained natively in a multimodal fashion through large-scale joint pretraining across vision and language, including proprietary systems such as GPT-4o and Gemini \cite{gpt-4o, gemini} as well as open-source models such as Kosmos-2, Molmo, and Transfusion \cite{kosmos-2, molmo-2, transfusion}. Such LMMs have demonstrated strong capabilities across a wide range of complex visual understanding and multimodal reasoning tasks, including visual question answering and compositional visual reasoning \cite{vqa,gqa,vqav2}, document and scene text understanding \cite{textvqa,docvqa,ocrvqa}, chart and infographic reasoning \cite{chartqa,infographicvqa,plotqa}, multimodal knowledge-intensive reasoning and large-scale multimodal evaluation benchmarks \cite{mmmu,mmbench,seedbench}, and real-world visual reasoning under noisy or unconstrained imagery \cite{vizwiz,naturalbench,realworldqa}.

\looseness=-1 Despite these advances, recent studies increasingly show that modern LMMs remain imperfect visual reasoners. Under cross-modal conflict, they can place disproportionate trust in textual inputs and follow misleading language cues over grounded visual evidence \cite{blindfaithtext}. They also remain prone to object hallucination and image-context reasoning failures, even when prompted on seemingly simple visual facts \cite{pope,hallusionbench}. Complementary analyses of their internal behavior further reveal that visual evidence is often weakly utilized: LMMs can exhibit \emph{visual attention sink} effects, rapidly diminishing contribution of visual tokens with depth, and highly sparse visual grounding behavior concentrated in only a few attention heads \cite{visualattentionsink,fastv,fewgroundheads}. More broadly, recent work on multimodal representation analysis suggests that these models continue to suffer from imperfect cross-modal alignment and incomplete modality integration within the LLM pathway \cite{alignvlm,mir}. Collectively, these findings suggest that, despite strong task-level performance, current LMM training still provides insufficient direct supervision to internal visual representations, leaving them vulnerable to text-dominant behaviors and brittle multimodal reasoning.

\begin{figure*}[t]
    \centering
    \includegraphics[width=0.78\textwidth]{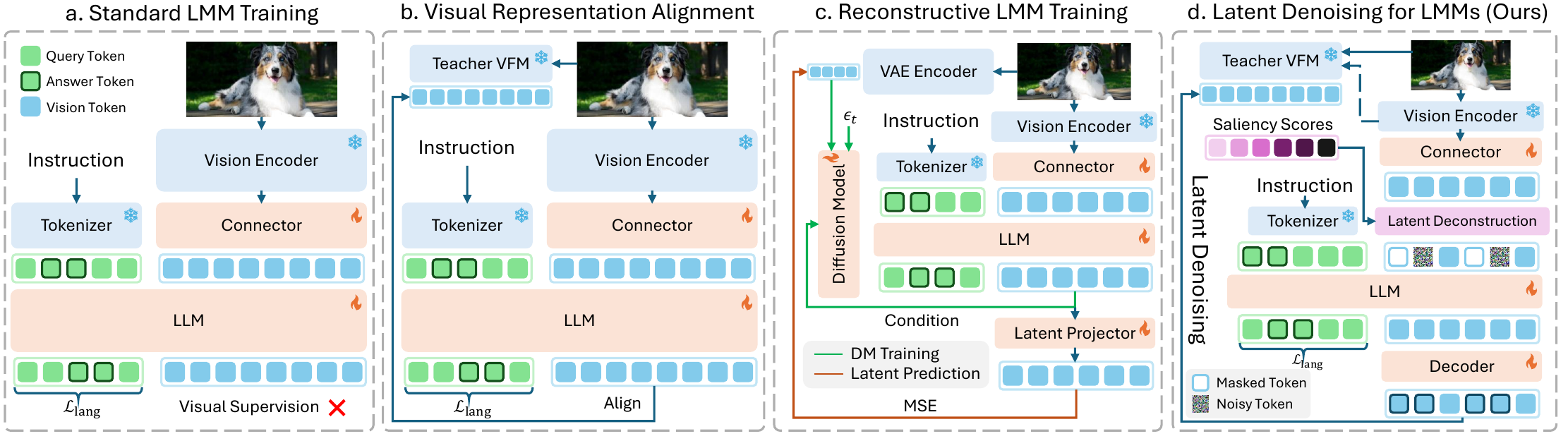}
    \caption{
    \textbf{Comparison of visual supervision paradigms for LMMs.}
    \textbf{(a)} Standard training: language loss only; visual tokens indirectly constrained.
    \textbf{(b)} Representation alignment: features aligned to external teachers \cite{basic,viral,vaco,finegrained}.
    \textbf{(c)} Reconstructive supervision: latent or diffusion-style reconstruction \cite{ross,laver,dsvlm}.
    \textbf{(d)} \emph{Ours}: saliency-guided corruption of projected visual tokens via Gaussian noising and masking, with recovery of clean teacher targets from intermediate hidden states.}
    \Description{Four side-by-side block diagrams comparing visual supervision paradigms in large multimodal models: standard language-loss-only training, representation alignment to external teachers, reconstructive supervision, and our latent denoising with saliency-guided corruption and mid-layer recovery.}
    \label{fig:framework_comparison}
\end{figure*}

\looseness=-1 Recent work has begun to address this limitation more directly. Some approaches first strengthen the visual backbone itself, for example by post-training vision encoders for better locality or aligning CLIP-like features to stronger vision-centric representations, thereby improving the visual inputs supplied to downstream LMMs while remaining largely external to end-to-end multimodal instruction tuning \cite{localityalignment,kernelclip}. Others inject explicit supervision into the multimodal stack by guiding projector outputs or internal visual tokens using refined shallow-layer embeddings, patch-level alignment, or alignment to one or more vision foundation models \cite{basic,finegrained,viral,vaco}. A third line introduces reconstructive visual supervision through latent reconstruction, masked image modeling, or diffusion-based objectives \cite{ross,laver,dsvlm}. Collectively, these studies make clear that direct visual supervision matters (Figure~\ref{fig:framework_comparison}); however, existing approaches remain fragmented across backbone refinement, feature alignment, and reconstructive supervision, and often rely on specialized teachers, auxiliary generators, or narrowly scoped supervision objectives. As a result, a more unified and efficient mechanism for refining internal visual representations within LMMs remains underexplored.

\looseness=-1 Motivated by this gap, we draw inspiration from a recent line of work suggesting that high-quality visual latent spaces need not be specific to either understanding or generation. On the generation side, semantically rich representation spaces have been shown to support strong diffusion-based image synthesis through representation autoencoders, while more recent unified multimodal pretraining results suggest that a shared visual representation can simultaneously benefit visual understanding and generation within a single model \cite{rae,scale-rae,transfusion}. In parallel, recent work on visual tokenizers shows that \emph{latent denoising}---recovering clean latent targets from corrupted latent inputs, e.g., under masking or Gaussian perturbations---can substantially improve the quality of learned visual tokens for downstream generation \cite{detok}. These developments motivate a simple question for large multimodal models: \emph{if latent denoising improves the quality of visual token spaces for generation, can it also serve as an effective form of supervision for the internal visual features used for multimodal understanding?}

\looseness=-1 Guided by this perspective, we propose a training-time latent denoising framework for large multimodal models (Figure~\ref{fig:method_overview}). Rather than relying on target-text supervision alone to shape visual representations, our method perturbs projected image tokens using a saliency-aware mixture of masking and Gaussian noising \cite{detok}, and trains the model to recover aligned teacher patch features from these corrupted tokens. The supervision is applied to an internal LMM layer, following recent observations that intermediate layers often preserve stronger visual signal than the final layer \cite{viral}, and is complemented by structure-preserving and discriminative objectives that retain intra-image geometry and patch specificity \cite{kernelclip,vaco}. Importantly, all auxiliary corruption and supervision are used only during training, so inference remains identical to the underlying LMM with no additional overhead.

\looseness=-1 We evaluate the proposed framework across three widely used open-source LMM variants spanning distinct vision encoders and model families, using 18 multimodal benchmarks covering visual question answering, OCR- and document-centric understanding, chart and structured visual reasoning, general multimodal reasoning, and robustness-oriented settings targeting real-world low-quality imagery and compositional consistency. Beyond standard evaluation, we adapt the common-corruptions paradigm of Hendrycks and Dietterich \cite{hendrycks2019benchmarking} to LMM assessment, introducing a deterministic, non-adversarial corruption protocol that applies noise, blur, weather, and digital corruptions---with multiple subtypes sampled within each family---directly to the evaluation images of these benchmark datasets \cite{vqarobustness}. Across these settings, the proposed framework consistently improves multimodal understanding, reasoning, and robustness over strong baselines, while maintaining higher accuracy and reduced degradation under corrupted-input evaluation. \emph{Our contributions are threefold:}
\begin{itemize}
    \item We introduce a training-time latent denoising framework for LMM instruction tuning that uses saliency-guided masking and Gaussian noising, teacher-guided feature recovery, \& structure-preserving objectives to refine internal visual representations.
    
    \item We develop a practical common-corruption evaluation protocol for LMMs, inspired by the ImageNet-C benchmark \cite{hendrycks2019benchmarking} and related visual robustness work in VQA \cite{vqarobustness}, enabling systematic non-adversarial evaluation under noise, blur, weather, and digital corruptions applied directly to benchmark evaluation images.
    
    \item We show that latent denoising consistently improves multimodal understanding, reasoning, and robustness across architectures, including under corrupted-input stress tests, while introducing no additional inference overhead.
\end{itemize}
\section{Related Works}
\label{sec:related}

\looseness=-1 \noindent \textbf{Large multimodal models.}
The dominant open-source LMM paradigm pairs a pretrained LLM with a vision encoder through a lightweight connector, followed by multimodal instruction tuning~\cite{flamingo,blip-2,instructblip,minigpt-4,mplug-owl,llama-adapter-v2,llava,llava-improved,llava-next,internvl,internvl-1.5,internvl-2.5,qwen-vl,qwen2-vl,qwen2.5-vl,qwen3-vl}. These models build on pretrained vision backbones such as CLIP and SigLIP~\cite{vit,clip,siglip,siglip-2}, while natively multimodal systems explore joint pretraining across modalities~\cite{gemini,gpt-4o,kosmos-2,molmo-2,transfusion}. Our work targets the open-source instruction-tuned setting where internal visual representation quality remains a central challenge.

\looseness=-1 \noindent \textbf{Visual supervision in LMMs.}
Recent work introduces direct visual supervision during LMM training through three broad strategies: backbone refinement via encoder-level alignment to stronger vision models~\cite{localityalignment,kernelclip}; feature alignment within the multimodal stack through projector refinement, patch-level supervision, or guidance from vision foundation models~\cite{basic,finegrained,viral,vaco}; and reconstructive supervision through latent reconstruction, masked modeling, or diffusion-based objectives~\cite{ross,laver,dsvlm}. Our work casts visual supervision as \emph{generalized latent denoising}, unifying continuous and discrete corruption with relational and contrastive losses.

\looseness=-1 \noindent \textbf{Denoising for visual representations.}
Corruption-and-recovery has long driven visual representation learning, from denoising autoencoders and masked image modeling~\cite{dae,mae} to diffusion-based generative models~\cite{ddpm,ldm}. Recent tokenizer work shows that combining masking and Gaussian corruption in latent space yields stronger visual tokenizers~\cite{detok}, while unified multimodal models demonstrate that strong visual latent spaces benefit both understanding and generation~\cite{tokenflow,transfusion,vqvae,vqgan,rae,scale-rae,textok}. We bring this latent denoising principle into multimodal instruction tuning for visual understanding.

\looseness=-1 \noindent \textbf{Corruption evaluation.}
The ImageNet-C common-corruptions paradigm~\cite{hendrycks2019benchmarking} has recently extended to multimodal settings through VQA robustness benchmarks~\cite{vqarobustness} and broader LMM robustness studies~\cite{rbench}, alongside adversarial robustness work~\cite{attackvlm,zeroshotadvvlm}. We adapt this non-adversarial corruption paradigm into a deterministic evaluation protocol for LMMs.

\section{Method}
\label{sec:method}

\subsection{Setup and Motivation}
\label{sec:method_setup}

\paragraph{Large multimodal models.}
\looseness=-1 Following the dominant recipe in open-source large multimodal models
\cite{flamingo,blip-2,instructblip,llava,qwen2-vl,internvl}, we consider a
decoder-only LMM built from three components: a frozen vision encoder
$E_{\boldsymbol{\phi}}$, a vision--language projector $P_{\boldsymbol{\psi}}$,
and an autoregressive language model $L_{\boldsymbol{\theta}}$. Given an input
image $\mathbf{X}\in\mathbb{R}^{H\times W\times 3}$, the vision encoder produces
a sequence of $S$ patch-level visual features
$\mathbf{V}=E_{\boldsymbol{\phi}}(\mathbf{X})
=[\mathbf{v}_1,\ldots,\mathbf{v}_S]$, where
$\mathbf{v}_i\in\mathbb{R}^{d_v}$, which are mapped into the language-model
embedding space through the projector
$\mathbf{Z}=P_{\boldsymbol{\psi}}(\mathbf{V})
=[\mathbf{z}_1,\ldots,\mathbf{z}_S]$, where
$\mathbf{z}_i\in\mathbb{R}^{d_h}$. Let
$\mathbf{Q}=[\mathbf{q}_1,\ldots,\mathbf{q}_M]\in\mathbb{R}^{M\times d_h}$
denote the tokenized user-query embeddings obtained from the LMM’s text
embedding layer. The LMM then processes the multimodal sequence
$[\mathbf{Z};\mathbf{Q}]$ and autoregressively models the target answer sequence
$\mathbf{A}=[\mathbf{a}_1,\ldots,\mathbf{a}_T]$ as
\begin{equation}
p_{\boldsymbol{\Theta}}(\mathbf{A}\mid \mathbf{Z},\mathbf{Q})
=
\prod_{t=1}^{T}
p_{\boldsymbol{\Theta}}
\!\left(
\mathbf{a}_t \mid \mathbf{a}_{<t}, \mathbf{Z}, \mathbf{Q}
\right),
\label{eq:setup_autoregressive}
\end{equation}
where $\boldsymbol{\Theta}=\{\boldsymbol{\phi},\boldsymbol{\psi},\boldsymbol{\theta}\}$. In
practice, $E_{\boldsymbol{\phi}}$ is typically a pretrained vision foundation
model such as a CLIP- or SigLIP-style encoder \cite{clip,siglip,siglip-2},
while $P_{\boldsymbol{\psi}}$ may be instantiated as an MLP projector, a
Q-Former, or a related cross-modal connector
\cite{blip-2,instructblip,llava}.

\paragraph{Standard LMM training.}
\looseness=-1 Most LMMs are trained in two stages \cite{llava,llava-improved,llava-next}. In
the first stage, often referred to as projector pretraining, the connector
$P_{\boldsymbol{\psi}}$ is optimized on image--text pairs while the vision
encoder and language model remain frozen. In the second stage, multimodal
instruction-tuning data is used to adapt the multimodal model for visual
instruction following, typically with the vision encoder still frozen and the
projector and language model trainable. In both stages, optimization is driven
by standard teacher-forced autoregressive supervision on the target answer
tokens. For notational simplicity, we write the multimodal training input as $[\mathbf{Z};\mathbf{U}]$, where $\mathbf{U}=[\mathbf{u}_1,\ldots,\mathbf{u}_U]$ denotes the tokenized text sequence containing the user instruction and target assistant response. Note that real training samples may be multi-turn.
The standard language
modeling loss is
\begin{equation}
\mathcal{L}_{\mathrm{lang}}
=
-
\sum_{t\in\mathcal{T}_{\mathrm{asst}}}
\log p_{\boldsymbol{\Theta}}
\!\left(
\mathbf{u}_t \mid \mathbf{u}_{<t}, \mathbf{Z}
\right),
\label{eq:setup_lang_loss}
\end{equation}
where $\mathcal{T}_{\mathrm{asst}}$ indexes the assistant positions in the
training sequence, i.e., the target answer tokens to be predicted. Under this
standard recipe, the trainable parameters are typically
$\{\boldsymbol{\psi},\boldsymbol{\theta}\}$, while the
visual backbone $\boldsymbol{\phi}$ remains frozen. As a result, internal visual
representations are supervised only indirectly through the final
text-generation objective, motivating more direct visual supervision during
multimodal instruction tuning.

\begin{figure*}[t]
    \centering
    \includegraphics[width=0.77\textwidth]{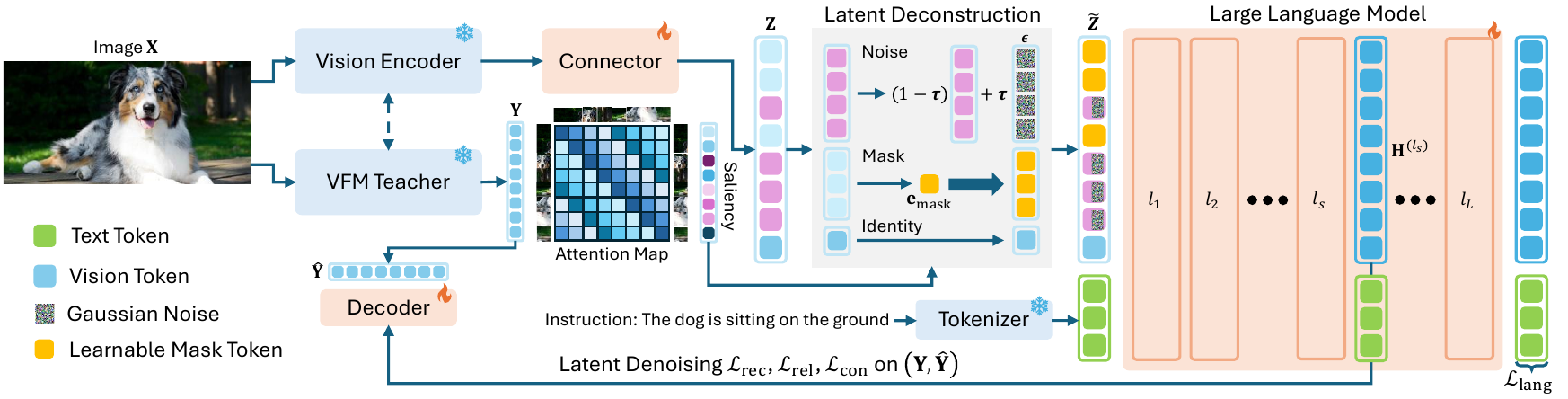}
    \caption{
    \textbf{Overview of latent denoising.}
    Projected visual tokens $\mathbf{Z}$ are corrupted via saliency-guided Gaussian noising and masking to yield $\tilde{\mathbf{Z}}$. The LMM processes corrupted tokens alongside the instruction, and a lightweight decoder recovers clean teacher targets $\hat{\mathbf{Y}}$ from an intermediate hidden layer. Training combines language modeling with reconstruction, relational, and contrastive denoising losses. All corruption and auxiliary decoding are training-only; inference is unchanged.
    }
    \Description{Pipeline diagram of latent denoising: projected visual tokens are corrupted by saliency-guided Gaussian noising and masking, processed by the language model with the instruction, and a lightweight decoder recovers clean teacher features from an intermediate hidden layer.}
    \label{fig:method_overview}
\end{figure*}

\paragraph{Latent denoising as supervision.}
\looseness=-1 We draw inspiration from recent work suggesting that latent denoising provides a
general mechanism for learning high-quality visual representations, and that
improvements in visual representation quality can benefit both understanding and
generation \cite{detok,rae,scale-rae,transfusion}. Latent
denoising consists of applying a corruption or \emph{deconstruction} process to
clean latent variables and then learning to recover aligned clean targets from
the corrupted latent inputs. Such corruption may take the form of discrete
information removal, as in masking-based objectives, continuous stochastic
perturbation, as in Gaussian denoising, or a combination of both
\cite{detok,mae,ddpm}. Formally, given clean latent visual tokens
$\mathbf{Z}$, we define corrupted latent tokens by sampling from a corruption
distribution
\begin{equation}
\tilde{\mathbf{Z}} \sim q(\tilde{\mathbf{Z}}\mid \mathbf{Z}),
\qquad\text{equivalently}\qquad
\tilde{\mathbf{Z}}=f_\mathcal{C}(\mathbf{Z}),
\label{eq:setup_corruption_operator}
\end{equation}
where $q(\tilde{\mathbf{Z}}\mid \mathbf{Z})$ denotes a latent corruption
distribution and $f_\mathcal{C}$ its corresponding corruption operator. To provide clean recovery targets, we
further define frozen teacher features
$\mathbf{Y}=F_{\boldsymbol{\zeta}}(\mathbf{X})
=[\mathbf{y}_1,\ldots,\mathbf{y}_S]$, where
$\mathbf{y}_i\in\mathbb{R}^{d_t}$ and $F_{\boldsymbol{\zeta}}$ is instantiated
in our experiments using the same frozen visual backbone as the front-end
encoder. Rather than supervising only the final generated text, our method
treats projected image tokens as a latent space, corrupts a subset of these
tokens, and trains the LMM to recover aligned clean teacher features from an
internal hidden layer. This reframes visual supervision in LMMs as a latent
corruption-and-recovery problem (Figure~\ref{fig:method_overview}), which is the central perspective of our method.

\subsection{Latent Corruption \& Recovery for LMMs}
\label{sec:method_denoise}

\paragraph{Gaussian noising.}
\looseness=-1 Given projected visual tokens
$\mathbf{Z}=[\mathbf{z}_1,\ldots,\mathbf{z}_S]$, we first define a soft
deconstruction process that perturbs a subset of tokens through Gaussian
noising. Let $\mathcal{N}\subseteq\{1,\ldots,S\}$ denote the set of indices
selected for Gaussian corruption, with $|\mathcal{N}|=K_N$. For each
$i\in\mathcal{N}$, we sample a token-specific corruption strength
$\tau_i\sim\mathrm{Unif}(0,\tau_{\max})$ and Gaussian noise
$\boldsymbol{\epsilon}_i\sim\mathcal{N}(\mathbf{0},\sigma^2\mathbf{I})$, and
define the corrupted token as
\begin{equation}
\tilde{\mathbf{z}}^{\,\mathrm{gauss}}_i
=
(1-\tau_i)\mathbf{z}_i+\tau_i\boldsymbol{\epsilon}_i,
\qquad i\in\mathcal{N}.
\label{eq:gaussian_noising}
\end{equation}
This operation preserves a controllable portion of the original visual signal
while forcing the LMM to recover aligned clean features from a degraded latent
input. For noised tokens, we additionally provide a lightweight
noise-conditioning signal indicating the magnitude of the applied corruption.

\paragraph{Masking.}
\looseness=-1 In parallel, we define hard deconstruction process via masking. Let
$\mathcal{M}\subseteq\{1,\ldots,S\}\setminus\mathcal{N}$ denote the set of
indices selected for masking, with $|\mathcal{M}|=K_M$. For each
$i\in\mathcal{M}$, the projected visual token is replaced with a learnable mask
embedding $\mathbf{e}_{\mathrm{mask}}\in\mathbb{R}^{d_h}$:
\begin{equation}
\tilde{\mathbf{z}}^{\,\mathrm{mask}}_i
=
\mathbf{e}_{\mathrm{mask}},
\qquad i\in\mathcal{M}.
\label{eq:mask_corruption}
\end{equation}
Masking removes token-level visual information entirely, creating a more severe
recovery problem than Gaussian noising. In our framework, masking therefore
plays role of hard latent deconstruction, complementing the softer
continuous perturbation in Eq.~\eqref{eq:gaussian_noising}.

\paragraph{Joint latent deconstruction.}
\looseness=-1 We combine the two deconstruction modes above into a single corruption process over the projected visual sequence. We write the
joint corrupted sequence as
\begin{equation}
\tilde{\mathbf{z}}_i=
\begin{cases}
(1-\tau_i)\mathbf{z}_i+\tau_i\boldsymbol{\epsilon}_i, & i\in\mathcal{N},\\[3pt]
\mathbf{e}_{\mathrm{mask}}, & i\in\mathcal{M},\\[3pt]
\mathbf{z}_i, & \text{otherwise},
\end{cases}
\label{eq:joint_corruption}
\end{equation}
and denote the full corrupted-token set by
$\mathcal{C}=\mathcal{N}\cup\mathcal{M}$. The sets $\mathcal{N}$ and
$\mathcal{M}$ are selected through a saliency-guided strategy derived from the
frozen visual encoder: visually salient patches are preferentially assigned to
Gaussian noising, while less salient patches are preferentially assigned to
masking. Intuitively, this produces a more balanced latent denoising problem.
Because the LMM remains a standard decoder-only model with causal attention, it
cannot rely on future visual positions to reconstruct heavily corrupted salient
content. We therefore apply \emph{soft} corruption to high-saliency tokens and
reserve \emph{hard} masking for lower-saliency tokens, typically with
$K_N > K_M$, so that the recovery task remains challenging but tractable. A
practical advantage of this design is that it requires no modification to the
causal masking structure of the underlying LMM, making the method naturally
compatible with standard decoder-only backbones.

\paragraph{Recovering aligned visual targets.}
\looseness=-1 The corrupted visual sequence $\tilde{\mathbf{Z}}=[\tilde{\mathbf{z}}_1,\ldots,\tilde{\mathbf{z}}_S]$
is concatenated with the tokenized query embeddings and processed by the LMM in
the standard autoregressive manner. To recover clean aligned targets, we extract
visual hidden states from a selected internal layer $\ell_s$ of the language
model. Let $\mathbf{h}_i^{(\ell_s)}\in\mathbb{R}^{d_h}$ denote the hidden state
at the $i$-th visual token position. We then attach a lightweight decoder
$D_{\boldsymbol{\chi}}:\mathbb{R}^{d_h}\rightarrow\mathbb{R}^{d_t}$ and decode
only the corrupted-token positions:
\begin{equation}
\hat{\mathbf{y}}_i
=
D_{\boldsymbol{\chi}}\!\left(\mathbf{h}_i^{(\ell_s)}\right),
\qquad i\in\mathcal{C}.
\label{eq:decode_internal}
\end{equation}
In practice, we apply this decoding from an intermediate layer of the LMM rather
than the final layer, motivated by recent evidence that visual signal is often
strongest in the middle of the network and progressively diluted at greater
depths \cite{viral,basic,fastv}. This choice leaves the final next-token prediction pathway unchanged while encouraging the LMM to maintain stronger and better-aligned internal visual features, thereby improving visual alignment for downstream multimodal reasoning.

\subsection{Training Objective}
\label{sec:method_objective}

\paragraph{Reconstruction loss.}
\looseness=-1 Given decoded predictions
$\hat{\mathbf{y}}_i=D_{\boldsymbol{\chi}}(\mathbf{h}_i^{(\ell_s)})$ for
corrupted token positions $i\in\mathcal{C}$, we first impose a normalized
reconstruction loss that aligns each prediction with its clean teacher target
$\mathbf{y}_i$:
\begin{equation}
\mathcal{L}_{\mathrm{rec}}
=
\frac{1}{|\mathcal{C}|}
\sum_{i\in\mathcal{C}}
\left\|
\mathrm{norm}(\hat{\mathbf{y}}_i)-
\mathrm{norm}(\mathbf{y}_i)
\right\|_2^2,
\label{eq:lrec}
\end{equation}
where $\mathrm{norm}(\mathbf{a})=\mathbf{a}/\|\mathbf{a}\|_2$. This gives direct
patch-level supervision to the internal visual representations used by the LMM,
while making the alignment invariant to feature scale
\cite{ross,basic,viral}.

\paragraph{Relational alignment loss.}
\looseness=-1 Pointwise recovery alone does not guarantee that decoded features preserve the
teacher's global visual structure. We therefore align the pairwise similarity
geometry of decoded and teacher features on the corrupted-token set
$\mathcal{C}$. Let
\begin{equation}
\mathbf{S}^{T}_{ij}=\cos(\mathbf{y}_i,\mathbf{y}_j),
\qquad
\mathbf{S}^{S}_{ij}=\cos(\hat{\mathbf{y}}_i,\hat{\mathbf{y}}_j),
\qquad i,j\in\mathcal{C},
\label{eq:sim_mats}
\end{equation}
denote the teacher and student similarity matrices. We then minimize a
row-wise KL divergence between the corresponding temperature-scaled similarity
distributions:
\begin{equation}
\mathcal{L}_{\mathrm{rel}}
=
\frac{1}{|\mathcal{C}|}
\sum_{i\in\mathcal{C}}
\mathrm{KL}
\Big(
\mathrm{softmax}(\mathbf{S}^{T}_{i:}/\tau_r)
\,\|\, 
\mathrm{softmax}(\mathbf{S}^{S}_{i:}/\tau_r)
\Big).
\label{eq:lrel}
\end{equation}
This encourages the recovered visual features to remain globally consistent with
the teacher representation rather than only matching it pointwise
\cite{kernelclip,rkd}.

\paragraph{Contrastive loss.}
\looseness=-1 Even with pointwise and relational alignment, decoded features may still become
weakly discriminative or collapse toward overly similar representations. To
encourage patch-level distinctiveness, we add an intra-image contrastive loss:
\begin{equation}
\mathcal{L}_{\mathrm{con}}
=
-\frac{1}{|\mathcal{C}|}
\sum_{i\in\mathcal{C}}
\log
\frac{
\exp\!\left(\cos(\hat{\mathbf{y}}_i,\mathbf{y}_i)/\tau_c\right)
}{
\sum_{j\in\mathcal{C}}
\exp\!\left(\cos(\hat{\mathbf{y}}_i,\mathbf{y}_j)/\tau_c\right)
}.
\label{eq:lcon}
\end{equation}
Here, each decoded corrupted token is treated as a positive pair with its own
teacher target and as a negative pair against the remaining corrupted-token
targets in the same image. This complements the relational loss by explicitly
discouraging degenerate solutions and sharpening patch-level discrimination
\cite{vaco,simclr}.

\paragraph{Overall objective.}
\looseness=-1 The final training objective combines the standard language modeling loss in
Eq.~\eqref{eq:setup_lang_loss} with the three visual supervision terms:
\begin{equation}
\mathcal{L}
=
\mathcal{L}_{\mathrm{lang}}
+
\lambda_{\mathrm{rec}}\mathcal{L}_{\mathrm{rec}}
+
\lambda_{\mathrm{rel}}\mathcal{L}_{\mathrm{rel}}
+
\lambda_{\mathrm{con}}\mathcal{L}_{\mathrm{con}}.
\label{eq:total_loss}
\end{equation}
All auxiliary losses are computed only on corrupted visual tokens. At inference
time, corruption is disabled and the auxiliary decoder is inactive, so the model
reduces to the original LMM forward pass with no additional runtime overhead.

\section{Experiments}
\label{sec:exp}

\looseness=-1 We evaluate the proposed method (Section \ref{sec:method}) across three LMM architectures spanning different vision encoders, language backbones, and training regimes. Our evaluation covers standard multimodal understanding benchmarks, robustness-oriented benchmarks, and a systematic corruption evaluation protocol. We further conduct ablation studies to analyze contributions of individual design choices.

\subsection{Setup}
\label{sec:exp_impl}

\looseness=-1 We instantiate latent denoising on three architectures: LLaVA-1.5~\cite{llava,llava-improved} with CLIP ViT-L/14@336~\cite{clip} and with SigLIP-SO400M/14@384~\cite{siglip}, both paired with Vicuna-7B-v1.5; and Qwen-2.5-VL-7B-Instruct~\cite{qwen2.5-vl}, post-tuned on LLaVA-665K. LLaVA variants follow standard two stage recipe (stage~1: projector pretraining on 558K, stage~2: instruction tuning on 665K with latent denoising enabled under a warmup--hold--decay schedule). Ablations use LoRA~\cite{lora} ($r\!=\!128$, $\alpha\!=\!256$). Full training details in Appendix~\ref{sec:appendix_training}. We evaluate on 18 benchmarks spanning visual QA~\cite{vqav2,gqa}, hallucination~\cite{pope,hallusionbench}, general reasoning~\cite{mme,mmbench,mmstar,mmmu}, science and math~\cite{scienceqa,mathvista}, text understanding~\cite{textvqa,ocrbench}, charts and diagrams~\cite{chartqa,ai2d}, and robustness~\cite{realworldqa,vizwiz,naturalbench,qbench}.

\subsection{Corruption Evaluation Protocol}
\label{sec:exp_corruption}

\looseness=-1 We adapt the ImageNet-C paradigm~\cite{hendrycks2019benchmarking,vqarobustness} into a deterministic corruption protocol for LMMs, applying corruptions from four families---noise, blur, weather, and digital---at severity levels 3 and~5. For each image, a corruption subtype is deterministically selected via a seeded hash, ensuring reproducibility. Crucially, our training-time corruption operates \emph{in the projected latent space}, while evaluation-time corruption is applied \emph{to input images}---the two are independent, testing whether improved internal representations transfer to robustness under input-level distribution shifts.

\noindent \textbf{Research Questions.} Our experiments are organized around four core questions:
\emph{(i)}~Does latent denoising improve multimodal understanding across architectures? (Sec.~\ref{sec:exp_main})
\emph{(ii)}~Does it improve robustness under corrupted visual inputs? (Sec.~\ref{sec:exp_robustness})
\emph{(iii)}~Does it produce measurably better internal visual representations? (Sec.~\ref{sec:exp_features})
\emph{(iv)}~How do individual design choices contribute to the overall framework? (Sec.~\ref{sec:exp_ablations})

\subsection{Main Results}
\label{sec:exp_main}

Table~\ref{tab:main_results} presents clean-evaluation results across three LMM architectures: LLaVA with CLIP ViT-L/14@336 (our primary setting), LLaVA with SigLIP-SO400M/14@384, and Qwen-2.5-VL-7B-Instruct. For each architecture, we compare a matched baseline trained under identical conditions without latent denoising.

\begin{table}[t]
\centering
\caption{\textbf{Main results across three LMM architectures.} B = Baseline, LD = Latent Denoising (ours). Best in each pair is \textbf{bolded}. NaturalBench reports G\textsubscript{ACC}.}
\label{tab:main_results}
\resizebox{0.94\columnwidth}{!}{%
\scriptsize
\setlength{\tabcolsep}{2.5pt}
\begin{tabular}{ll|ccc|ccc|ccc}
\toprule
& & \multicolumn{3}{c|}{\textbf{LLaVA+CLIP}} & \multicolumn{3}{c|}{\textbf{LLaVA+SigLIP}} & \multicolumn{3}{c}{\textbf{Qwen-2.5-VL}} \\
& & B & LD & $\Delta$ & B & LD & $\Delta$ & B & LD & $\Delta$ \\
\midrule
\multirow{2}{*}{\rotatebox{90}{\tiny VQA}}
& VQAv2 & 76.7 & \textbf{77.3} & \cmark{+0.6} & 76.3 & \textbf{78.6} & \cmark{+2.3} & 82.2 & \textbf{83.4} & \cmark{+1.2} \\
& GQA & 61.9 & \textbf{63.5} & \cmark{+1.6} & 62.1 & \textbf{64.3} & \cmark{+2.2} & 60.2 & \textbf{64.6} & \cmark{+4.4} \\
\midrule
\multirow{2}{*}{\rotatebox{90}{\tiny Hall.}}
& POPE & 87.0 & \textbf{87.7} & \cmark{+0.7} & 85.4 & \textbf{87.2} & \cmark{+1.8} & 87.2 & \textbf{88.1} & \cmark{+0.9} \\
& HallBench & 27.6 & \textbf{29.6} & \cmark{+2.0} & 28.3 & \textbf{30.7} & \cmark{+2.4} & 52.9 & \textbf{54.7} & \cmark{+1.8} \\
\midrule
\multirow{4}{*}{\rotatebox{90}{\tiny General}}
& MME & 1879 & 1846 & \xmark{-33} & 1708 & \textbf{1733} & \cmark{+25} & \textbf{2279} & 2265 & \xmark{-15} \\
& MMBench & 64.0 & \textbf{65.7} & \cmark{+1.7} & 64.5 & \textbf{68.7} & \cmark{+4.2} & 83.1 & \textbf{84.5} & \cmark{+1.4} \\
& MMStar & 33.7 & \textbf{36.8} & \cmark{+3.1} & 36.9 & \textbf{38.8} & \cmark{+1.9} & 61.3 & \textbf{62.5} & \cmark{+1.2} \\
& MMMU & 36.1 & \textbf{36.2} & \cmark{+0.1} & 34.6 & \textbf{36.1} & \cmark{+1.5} & 50.4 & \textbf{52.3} & \cmark{+1.9} \\
\midrule
\multirow{2}{*}{\rotatebox{90}{\tiny Sci.}}
& SQA-IMG & \textbf{69.5} & 69.2 & \xmark{-0.3} & 70.6 & \textbf{72.4} & \cmark{+1.8} & 87.3 & \textbf{88.5} & \cmark{+1.2} \\
& MathVista & 25.6 & \textbf{27.6} & \cmark{+2.0} & 26.4 & \textbf{29.0} & \cmark{+2.6} & 68.2 & \textbf{70.3} & \cmark{+2.1} \\
\midrule
\multirow{2}{*}{\rotatebox{90}{\tiny Text}}
& TextVQA & 46.1 & \textbf{46.7} & \cmark{+0.6} & 49.6 & \textbf{54.1} & \cmark{+4.5} & 81.8 & \textbf{83.7} & \cmark{+1.9} \\
& OCRBench & 31.4 & \textbf{32.1} & \cmark{+0.7} & 32.2 & \textbf{34.8} & \cmark{+2.6} & 82.3 & \textbf{84.2} & \cmark{+1.9} \\
\midrule
\multirow{2}{*}{\rotatebox{90}{\tiny Chart}}
& ChartQA & 18.1 & \textbf{18.3} & \cmark{+0.2} & 15.9 & \textbf{16.9} & \cmark{+1.0} & 83.0 & \textbf{85.4} & \cmark{+2.4} \\
& AI2D & 55.5 & \textbf{57.2} & \cmark{+1.7} & 54.8 & \textbf{57.1} & \cmark{+2.3} & 83.9 & \textbf{85.4} & \cmark{+1.5} \\
\midrule
\multirow{4}{*}{\rotatebox{90}{\tiny Robust.}}
& NatBench & 13.3 & \textbf{15.7} & \cmark{+2.4} & 14.8 & \textbf{15.7} & \cmark{+0.9} & 31.3 & \textbf{33.5} & \cmark{+2.2} \\
& RWQA & 56.1 & 56.1 & {0.0} & 53.0 & \textbf{54.1} & \cmark{+1.1} & 68.7 & \textbf{72.9} & \cmark{+4.2} \\
& VizWiz & 54.0 & \textbf{54.8} & \cmark{+0.8} & 54.4 & \textbf{57.8} & \cmark{+3.4} & 68.9 & \textbf{70.9} & \cmark{+2.0} \\
& Q-Bench & 58.7 & \textbf{59.6} & \cmark{+0.9} & 57.4 & \textbf{59.1} & \cmark{+1.7} & 73.9 & \textbf{75.9} & \cmark{+2.0} \\
\bottomrule
\end{tabular}%
}
\end{table}

\noindent \textbf{Consistent improvement across architectures.}
Latent denoising yields positive improvements on the large majority of benchmarks across all three architectures: 15 out of 18 for LLaVA+CLIP, 18 out of 18 for LLaVA+SigLIP, and 17 out of 18 for Qwen-2.5-VL. Notably, the method improves a strong, already-instruction-tuned Qwen model that was not trained from scratch with our framework, demonstrating that latent denoising can refine visual representations even during post-tuning of mature models.

\noindent \textbf{Largest gains on visual reasoning and grounding.}
The most substantial improvements appear on benchmarks that demand fine-grained visual reasoning. GQA, which tests compositional spatial reasoning, improves by +1.6, +2.2, and +4.4 across the three architectures. MMStar, which requires holistic multimodal understanding, improves by +3.1, +1.9, and +1.2. NaturalBench group accuracy, a strict compositional consistency metric, improves by +2.4, +0.9, and +2.2. These gains are consistent with our hypothesis that training the LMM to recover aligned visual features from corrupted latent inputs strengthens the internal visual representations used for downstream reasoning.

\noindent \textbf{Robustness benchmarks benefit broadly.}
Even without any corruption applied at evaluation time, robustness-oriented benchmarks that feature naturally challenging imagery show clear improvements. VizWiz (+0.8 to +3.4 across architectures), Q-Bench (+0.9 to +2.0), and RealWorldQA (+1.1 to +4.2 on SigLIP and Qwen) all benefit, suggesting that stronger internal visual representations translate to better handling of low-quality visual inputs.

\noindent \textbf{MME as an outlier.}
The combined MME score (perception + cognition) is the only metric that shows a consistent slight decline on two of three architectures (LLaVA+CLIP: $-32.7$, Qwen: $-14.6$), while improving on SigLIP (+24.4). We note that MME uses a binary yes/no format with perception subtasks that emphasize existence, count, and position; such coarse judgments may be less sensitive to the fine-grained visual alignment improvements that latent denoising provides.

\subsection{Corruption Robustness}
\label{sec:exp_robustness}

\looseness=-1 We now evaluate whether the improved visual representations produced by latent denoising translate to robustness under corrupted visual inputs. Following the protocol described in Sec.~\ref{sec:exp_corruption}, we apply four corruption families---noise, blur, weather, and digital---at severity levels 3 and~5 to the images of each evaluation benchmark, and compare baseline and latent denoising models under identical corrupted conditions.

\noindent \textbf{Corruption results.}
Tables~\ref{tab:corruption_clip} and~\ref{tab:corruption_qwen} report severity-5 results. Latent denoising improves all four corruption families on all 11 benchmarks for both architectures. Notably, MME---which declines slightly under clean evaluation---turns positive under all corruption conditions, suggesting benefit under degraded visual inputs.

\begin{table}[t]
\centering
\caption{\textbf{Corruption robustness at severity~5 (LLaVA+CLIP).} B~=~Baseline, LD~=~Latent Denoising (ours).}
\label{tab:corruption_clip}
\resizebox{0.94\columnwidth}{!}{%
\scriptsize
\setlength{\tabcolsep}{2pt}
\begin{tabular}{l|ccc|ccc|ccc|ccc}
\toprule
& \multicolumn{3}{c|}{\textbf{Noise}} & \multicolumn{3}{c|}{\textbf{Blur}} & \multicolumn{3}{c|}{\textbf{Weather}} & \multicolumn{3}{c}{\textbf{Digital}} \\
& B & LD & $\Delta$ & B & LD & $\Delta$ & B & LD & $\Delta$ & B & LD & $\Delta$ \\
\midrule
VQAv2 & 69.4 & \textbf{71.1} & \cmark{+1.7} & 67.8 & \textbf{69.3} & \cmark{+1.5} & 71.4 & \textbf{73.3} & \cmark{+1.9} & 70.9 & \textbf{73.0} & \cmark{+2.1} \\
GQA & 57.4 & \textbf{60.4} & \cmark{+3.0} & 56.4 & \textbf{59.7} & \cmark{+3.3} & 59.0 & \textbf{62.0} & \cmark{+3.0} & 58.4 & \textbf{61.6} & \cmark{+3.2} \\
POPE & 80.7 & \textbf{82.8} & \cmark{+2.1} & 77.8 & \textbf{80.2} & \cmark{+2.4} & 82.5 & \textbf{84.7} & \cmark{+2.2} & 81.8 & \textbf{83.9} & \cmark{+2.1} \\
MME & 1712 & \textbf{1728} & \cmark{+16} & 1678 & \textbf{1699} & \cmark{+21} & 1679 & \textbf{1705} & \cmark{+26} & 1702 & \textbf{1729} & \cmark{+28} \\
MMB & 54.2 & \textbf{56.3} & \cmark{+2.1} & 53.5 & \textbf{55.3} & \cmark{+1.8} & 57.3 & \textbf{60.2} & \cmark{+2.9} & 56.7 & \textbf{58.7} & \cmark{+2.0} \\
MMStar & 29.5 & \textbf{35.3} & \cmark{+5.8} & 30.8 & \textbf{36.1} & \cmark{+5.3} & 32.5 & \textbf{36.8} & \cmark{+4.3} & 30.6 & \textbf{35.8} & \cmark{+5.2} \\
MMMU & 35.5 & \textbf{37.3} & \cmark{+1.8} & 34.6 & \textbf{35.5} & \cmark{+0.9} & 33.6 & \textbf{36.7} & \cmark{+3.1} & 35.2 & \textbf{36.9} & \cmark{+1.7} \\
SQA & 68.2 & \textbf{69.0} & \cmark{+0.8} & 67.8 & \textbf{69.4} & \cmark{+1.6} & 67.6 & \textbf{68.9} & \cmark{+1.3} & 68.1 & \textbf{69.5} & \cmark{+1.4} \\
TxtVQA & 35.0 & \textbf{36.5} & \cmark{+1.5} & 24.8 & \textbf{26.5} & \cmark{+1.7} & 38.1 & \textbf{40.3} & \cmark{+2.2} & 37.5 & \textbf{38.7} & \cmark{+1.2} \\
OCR & 15.8 & \textbf{17.7} & \cmark{+1.9} & 12.5 & \textbf{14.2} & \cmark{+1.7} & 26.1 & \textbf{28.5} & \cmark{+2.4} & 18.9 & \textbf{21.2} & \cmark{+2.3} \\
Chart & 13.2 & \textbf{15.3} & \cmark{+2.1} & 12.4 & \textbf{13.8} & \cmark{+1.4} & 14.8 & \textbf{16.4} & \cmark{+1.6} & 14.1 & \textbf{15.6} & \cmark{+1.5} \\
\bottomrule
\end{tabular}%
}
\end{table}

\begin{table}[t]
\centering
\caption{\textbf{Corruption robustness at severity~5 (Qwen-2.5-VL).} B~=~Baseline, LD~=~Latent Denoising (ours).}
\label{tab:corruption_qwen}
\resizebox{0.94\columnwidth}{!}{%
\scriptsize
\setlength{\tabcolsep}{2pt}
\begin{tabular}{l|ccc|ccc|ccc|ccc}
\toprule
& \multicolumn{3}{c|}{\textbf{Noise}} & \multicolumn{3}{c|}{\textbf{Blur}} & \multicolumn{3}{c|}{\textbf{Weather}} & \multicolumn{3}{c}{\textbf{Digital}} \\
& B & LD & $\Delta$ & B & LD & $\Delta$ & B & LD & $\Delta$ & B & LD & $\Delta$ \\
\midrule
VQAv2 & 58.8 & \textbf{60.6} & \cmark{+1.8} & 65.7 & \textbf{67.6} & \cmark{+1.9} & 73.3 & \textbf{75.4} & \cmark{+2.1} & 69.9 & \textbf{72.4} & \cmark{+2.5} \\
GQA & 47.3 & \textbf{53.7} & \cmark{+6.4} & 51.1 & \textbf{57.9} & \cmark{+6.8} & 54.7 & \textbf{62.3} & \cmark{+7.6} & 52.6 & \textbf{61.0} & \cmark{+8.4} \\
POPE & 72.9 & \textbf{75.1} & \cmark{+2.2} & 75.7 & \textbf{79.0} & \cmark{+3.3} & 81.5 & \textbf{84.3} & \cmark{+2.8} & 77.5 & \textbf{80.4} & \cmark{+2.9} \\
MME & 1785 & \textbf{1807} & \cmark{+22} & 1754 & \textbf{1778} & \cmark{+25} & 2079 & \textbf{2094} & \cmark{+15} & 2023 & \textbf{2054} & \cmark{+31} \\
MMB & 54.7 & \textbf{57.7} & \cmark{+3.0} & 60.2 & \textbf{62.5} & \cmark{+2.3} & 69.1 & \textbf{71.1} & \cmark{+2.0} & 63.5 & \textbf{67.4} & \cmark{+3.9} \\
MMStar & 44.8 & \textbf{47.6} & \cmark{+2.8} & 42.0 & \textbf{42.7} & \cmark{+0.7} & 53.6 & \textbf{56.7} & \cmark{+3.1} & 50.3 & \textbf{51.8} & \cmark{+1.5} \\
MMMU & 45.5 & \textbf{50.0} & \cmark{+4.5} & 43.7 & \textbf{46.7} & \cmark{+3.0} & 47.1 & \textbf{51.0} & \cmark{+3.9} & 46.4 & \textbf{48.7} & \cmark{+2.3} \\
SQA & 77.3 & \textbf{78.8} & \cmark{+1.5} & 77.4 & \textbf{80.0} & \cmark{+2.6} & 81.5 & \textbf{83.6} & \cmark{+2.1} & 81.7 & \textbf{84.2} & \cmark{+2.5} \\
TxtVQA & 34.1 & \textbf{36.5} & \cmark{+2.4} & 33.9 & \textbf{35.9} & \cmark{+2.0} & 70.9 & \textbf{73.0} & \cmark{+2.1} & 58.9 & \textbf{60.5} & \cmark{+1.6} \\
OCR & 31.7 & \textbf{33.7} & \cmark{+2.0} & 19.9 & \textbf{22.3} & \cmark{+2.4} & 68.6 & \textbf{71.8} & \cmark{+3.2} & 47.0 & \textbf{50.1} & \cmark{+3.1} \\
Chart & 23.4 & \textbf{26.1} & \cmark{+2.7} & 22.7 & \textbf{24.6} & \cmark{+1.9} & 62.9 & \textbf{66.0} & \cmark{+3.1} & 45.2 & \textbf{48.1} & \cmark{+2.9} \\
\bottomrule
\end{tabular}%
}
\end{table}

\begin{figure}[t]
    \centering
    \includegraphics[width=0.77\linewidth]{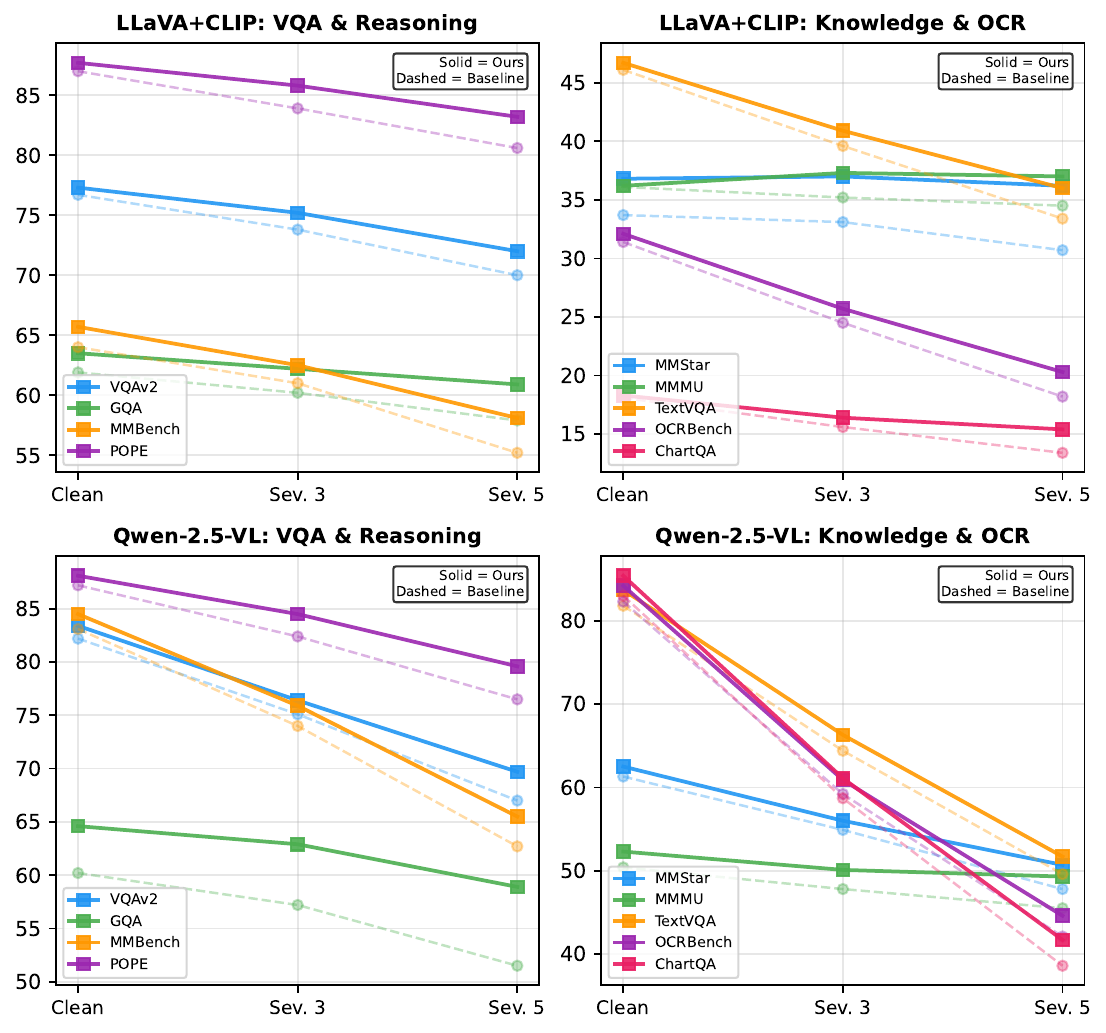}
    \caption{\textbf{Performance degradation under increasing corruption severity.} Clean $\to$ sev.~3 $\to$ sev.~5 for baseline (dashed) and latent denoising (solid) on LLaVA+CLIP (top) and Qwen-2.5-VL (bottom). The gap widens at higher severity, indicating reduced degradation.}
    \Description{Line charts of benchmark performance from clean to severity 3 to severity 5 corruption for baseline and latent denoising on LLaVA plus CLIP and Qwen 2.5 VL, showing smaller degradation and a widening gap for latent denoising.}
    \label{fig:corruption_degradation}
\end{figure}

\noindent \textbf{Degradation under increasing severity.}
Figure~\ref{fig:corruption_degradation} shows performance trajectories from clean to severity~3 to severity~5 on LLaVA+CLIP and Qwen-2.5-VL. On both architectures, models trained with latent denoising show consistently smaller drops from clean to corrupted performance, with the gap between baseline and our method widening as corruption severity increases. For example, on GQA under severity-5 noise (LLaVA+CLIP), the baseline drops from 61.9 to 57.4 ($-4.5$), while our model drops from 63.5 to 60.4 ($-3.1$), a meaningfully smaller degradation. On MMStar, the baseline drops by $-4.2$ while our model drops by only $-1.5$, and our corrupted performance (35.3) exceeds the baseline's \emph{clean} performance (33.7).

\begin{figure}[t]
    \centering
    \includegraphics[width=0.88\linewidth]{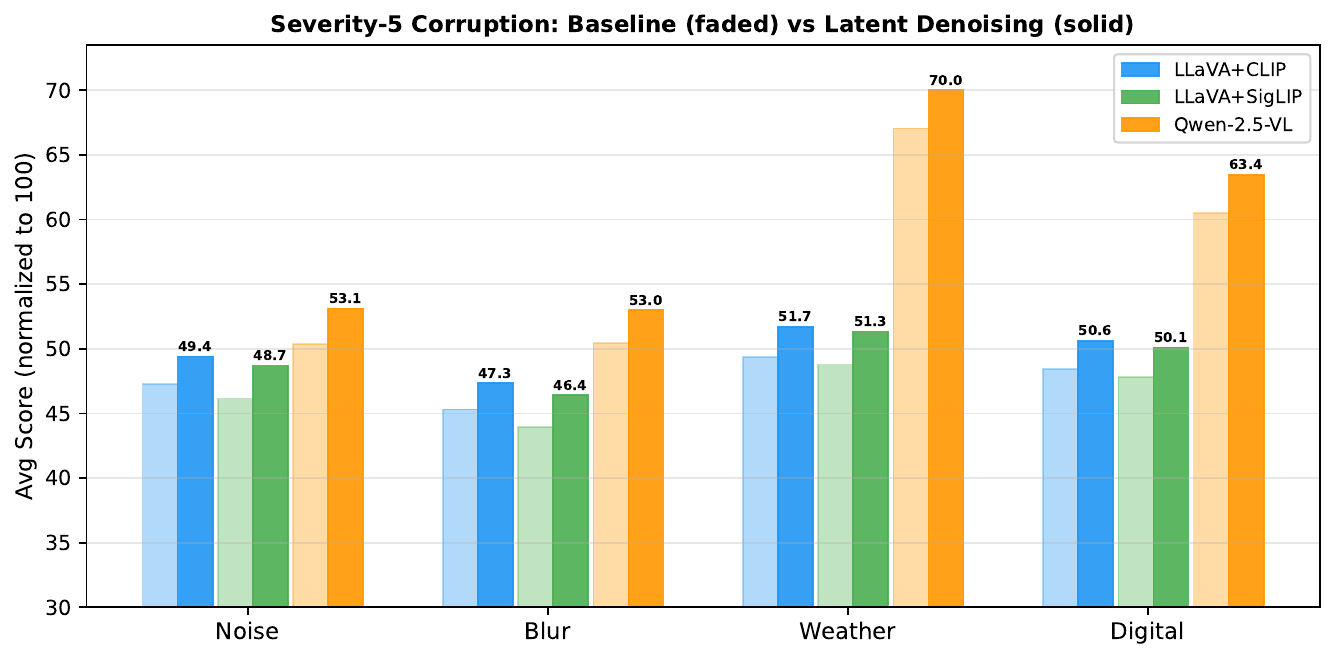}
    \caption{\textbf{Average normalized performance under severity-5 corruption.} Baseline (faded) vs.\ latent denoising (solid) across all 11 benchmarks (MME normalized to 100). All three architectures show consistent improvements across all four corruption families.}
    \Description{Grouped bar chart of average normalized severity 5 performance across noise, blur, weather, and digital corruption families for three architectures, with latent denoising bars consistently above baseline bars.}
    \label{fig:corruption_abs}
\end{figure}

\noindent \textbf{Cross-architecture consistency.}
Figure~\ref{fig:corruption_abs} confirms that the robustness benefit generalizes: latent denoising outperforms the baseline on every architecture--corruption combination at severity~5, with particularly pronounced gains on Qwen-2.5-VL.

\subsection{Analyzing Visual Feature Alignment}
\label{sec:exp_features}

Beyond task-level evaluation, we analyze whether latent denoising produces measurably better internal visual representations within the LLM. We extract mean-pooled visual hidden states from every layer of the 32-layer Vicuna backbone on a 5{,}000-image subset of ImageNet-1K validation, using the LLaVA+CLIP baseline and latent denoising models.

\begin{figure}[t]
    \centering
    \includegraphics[width=0.81\columnwidth]{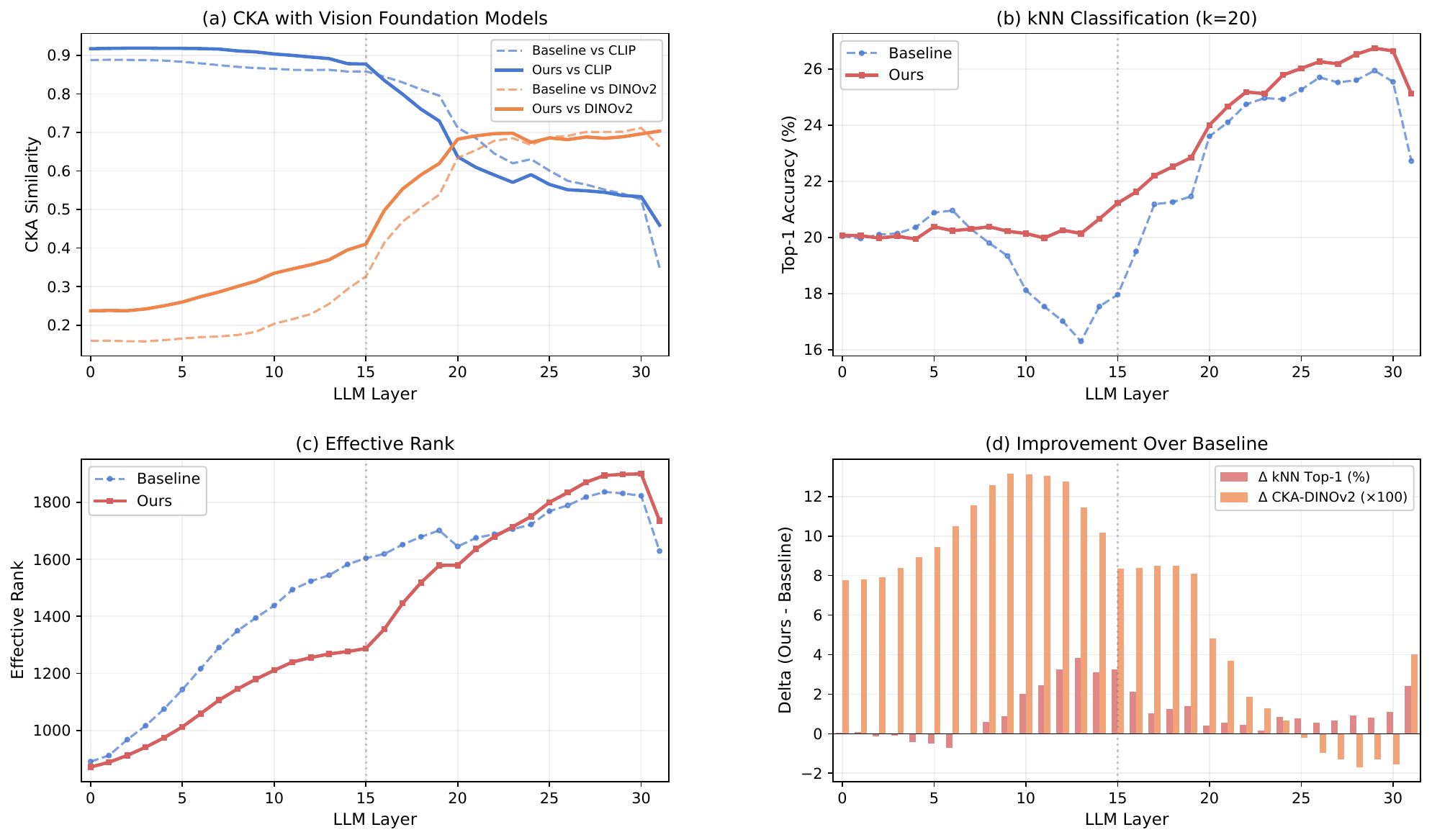}
    \caption{\textbf{Internal visual feature analysis across LLM layers} (LLaVA+CLIP, 5K ImageNet val). \textbf{(a)}~CKA with CLIP and DINOv2. \textbf{(b)}~$k$-NN accuracy. \textbf{(c)}~Effective rank. \textbf{(d)}~Improvement over baseline. Dashed line marks the supervised layer~15.}
    \Description{Four panels of layer-wise probing curves: CKA with CLIP, CKA with DINOv2, k nearest neighbor accuracy, and effective rank across language model layers, each showing latent denoising above baseline.}
    \label{fig:feature_analysis}
\end{figure}

\noindent \textbf{CKA with reference models.}
We compute linear centered kernel alignment (CKA)~\cite{cka} between each LLM layer and two frozen reference models: the same CLIP ViT-L/14 encoder used as the LMM's vision backbone, and DINOv2 ViT-L/14, a self-supervised vision model. Figure~\ref{fig:feature_analysis}(a) shows both: latent denoising maintains higher CKA with CLIP at early-to-mid layers (layers 0--15, +0.02 to +0.04) and dramatically improves final-layer alignment (+0.11 at layer~31), indicating that the LLM better preserves the visual information it receives from the encoder. CKA with DINOv2 reveals an even stronger effect: improvements of +0.08 to +0.13 at layers 0--15, peaking at +0.13 at layer~9. Since the denoising target is CLIP (not DINOv2), this cross-model alignment suggests that the denoising objective discovers structural improvements in the visual representation that generalize beyond the specific teacher.

\begin{figure}[t]
    \centering
    \includegraphics[width=0.81\linewidth]{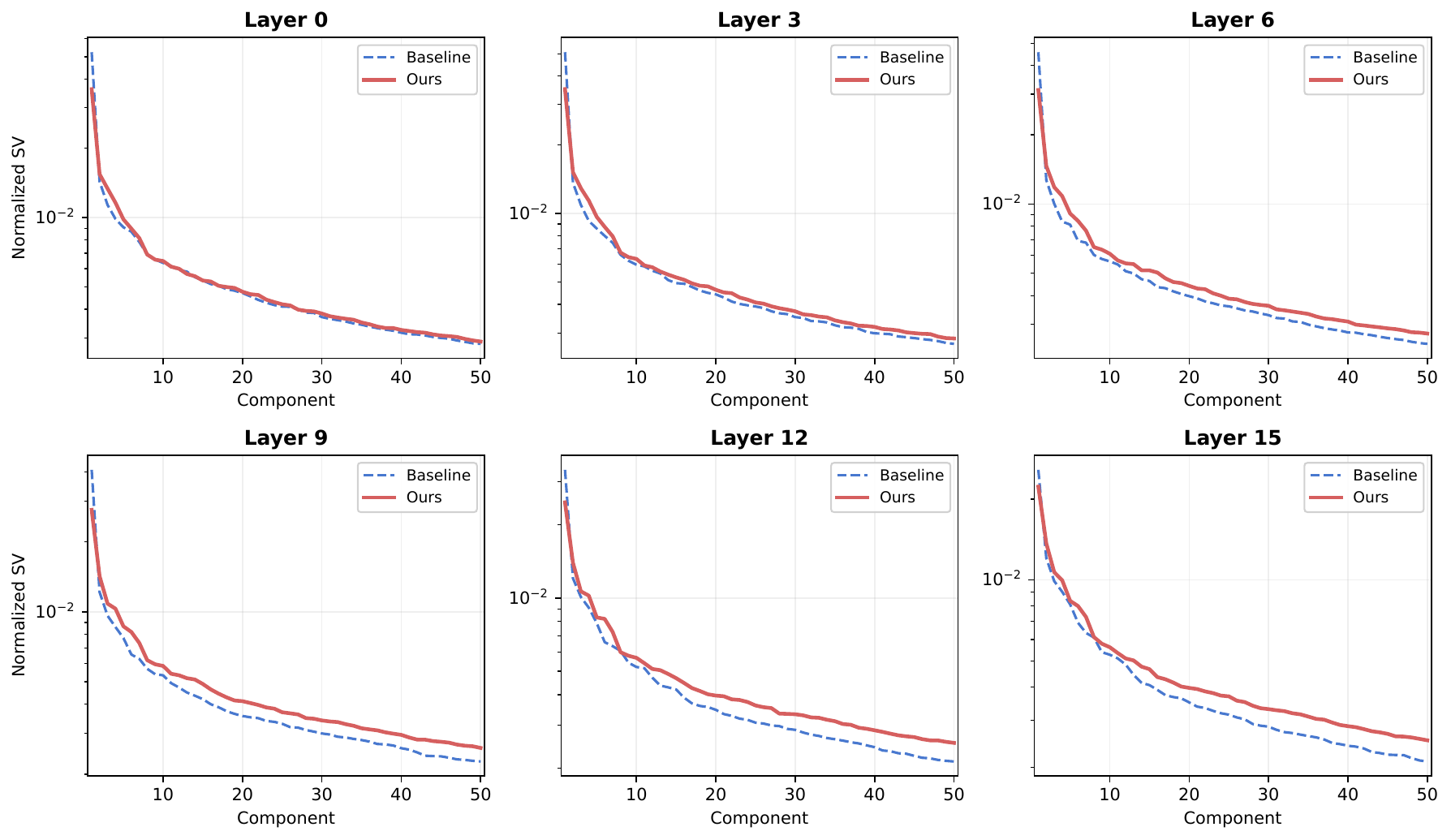}
    \caption{\textbf{Singular value spectrum at selected LLM layers.} Normalized singular value decay for visual features at layers 0, 3, 6, 9, 12, and 15. Latent denoising produces slower spectral decay at mid-layers, indicating higher intrinsic dimensionality and more expressive visual representations.}
    \Description{Normalized singular value spectra of visual features at selected language model layers, with latent denoising showing slower spectral decay than baseline at mid layers.}
    \label{fig:sv_spectrum}
\end{figure}

\noindent \textbf{kNN classification.}
Following standard practice for evaluating internal visual representations~\cite{knnprobe,clip}, we assess semantic discriminability via $k$-nearest-neighbor classification ($k\!=\!20$, cosine similarity, 5-fold cross-validation) on ImageNet-1K validation images. Figure~\ref{fig:feature_analysis}(b) shows top-1 accuracy per layer. The baseline exhibits a characteristic mid-network dip: features become progressively less discriminative from layers 5--13, recovering only at deeper layers as language-oriented representations emerge. Latent denoising eliminates this collapse. The largest gains occur at layers 10--15, centered on the supervised layer (layer~16), where our model improves by up to +3.8\% absolute. Critically, these gains persist through the final layer (+2.4\%), demonstrating that the supervision propagates beyond its direct point of application. The singular value spectrum (Figure~\ref{fig:sv_spectrum}) corroborates this: latent denoising produces slower spectral decay at mid-layers, indicating higher intrinsic dimensionality~\cite{effectiverank} and more expressive visual features.

\begin{figure}
    \centering
    \includegraphics[width=0.94\columnwidth,trim=0 18 0 18,clip]{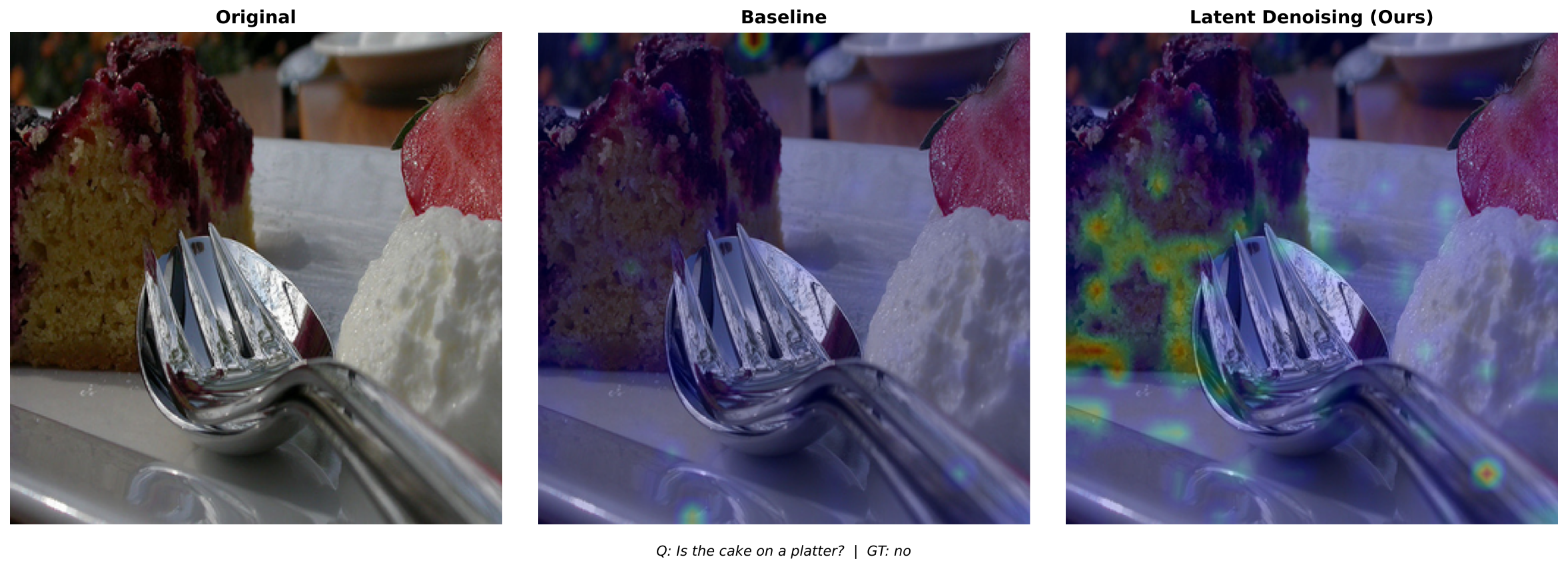}\\[-2pt]
    {\scriptsize\itshape Q: Is the cake on a platter? $\mid$ GT: no}\\[3pt]
    \includegraphics[width=0.94\columnwidth,trim=0 18 0 18,clip]{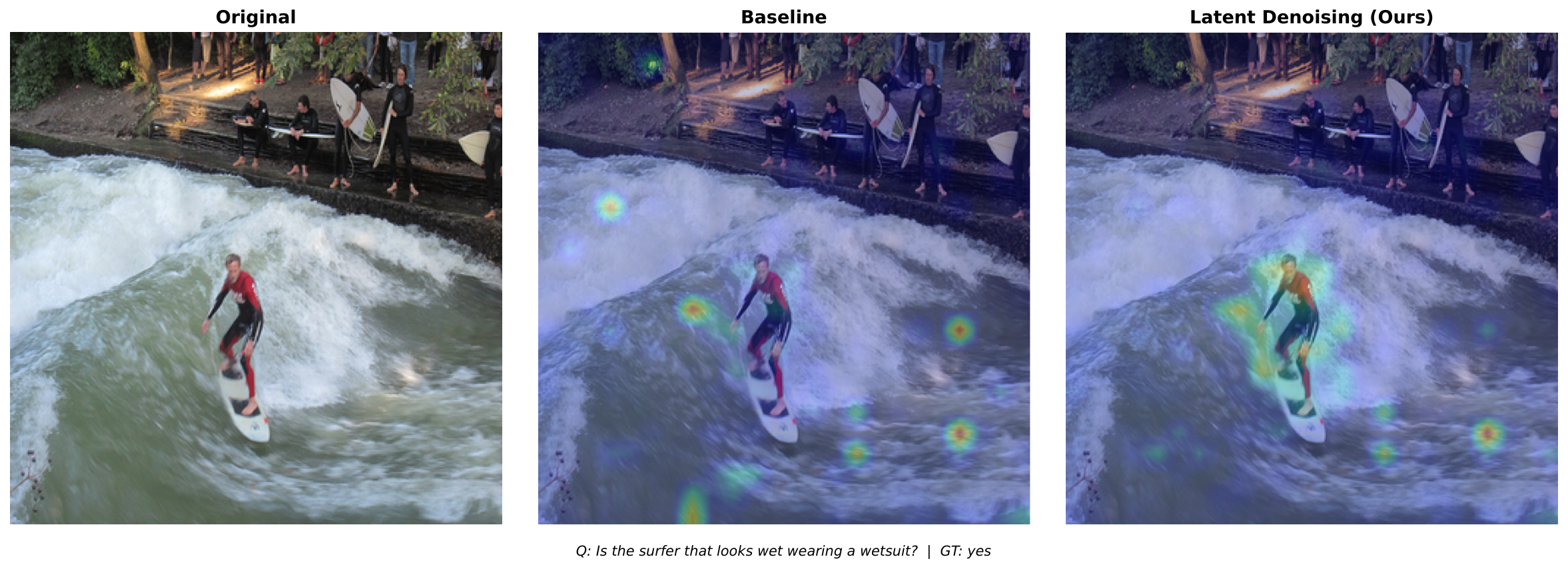}\\[-2pt]
    {\scriptsize\itshape Q: Is the surfer that looks wet wearing a wetsuit? $\mid$ GT: yes}
    \caption{\textbf{Cross-attention heatmaps on GQA} (layer~15). Each row shows, left to right: original image, baseline attention, and latent denoising attention. Our model produces more spatially grounded attention on question-relevant regions (cake/platter, wetsuit). Average attention entropy: ours 5.77 vs.\ baseline 5.49.}
    \Description{Cross-attention heatmaps at layer 15 for two GQA examples, showing original image, baseline attention, and latent denoising attention; latent denoising attends more precisely to question-relevant regions.}
    \label{fig:attention_grid}
\end{figure}

\noindent \textbf{Cross-attention visualization.}
Figure~\ref{fig:attention_grid} shows cross-attention heatmaps from layer~15 on GQA examples. Latent denoising produces more spatially focused attention on question-relevant regions, with higher average entropy (5.77 vs.\ 5.49) indicating engagement of multiple relevant patches rather than collapse onto attention sinks.

\subsection{Properties of Latent Denoising \& Ablations}
\label{sec:exp_ablations}

We ablate key design choices using LoRA fine-tuning on the CLIP+Vicuna configuration (see Sec.~\ref{sec:exp_impl}). All ablation results are normalized relative to the full latent denoising model to isolate the effect of each design choice.

\begin{figure}[t]
    \centering
    \includegraphics[width=0.88\columnwidth]{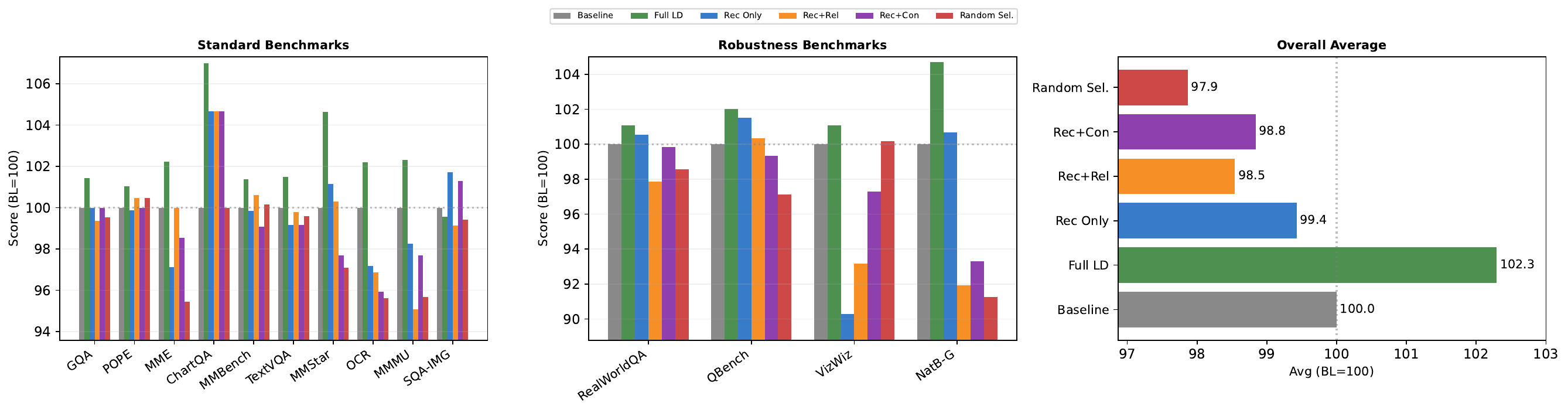}
    \caption{\textbf{Loss component ablation} (Baseline = 100). Full latent denoising with all three losses and saliency-guided selection achieves the best overall performance. Removing any loss or switching to random selection reduces gains.}
    \Description{Bar chart of loss component ablation normalized to full latent denoising, showing that removing any of the three losses or using random token selection reduces performance.}
    \label{fig:ablation_loss}
\end{figure}

\noindent \textbf{Loss components.}
Figure~\ref{fig:ablation_loss} compares the full three-loss objective against variants using subsets of losses and random (non-saliency) token selection, normalized to the LoRA baseline. The full model (all three losses with saliency-guided selection) achieves the highest overall performance. Removing any single loss component reduces performance, with the contrastive loss ($\mathcal{L}_\mathrm{con}$) and relational loss ($\mathcal{L}_\mathrm{rel}$) contributing most to MME and robustness benchmarks, respectively. Random token selection substantially degrades MME perception (1388.9 vs.\ 1489.3), confirming that saliency-guided corruption selection is important for preserving perceptual capabilities.

\noindent \textbf{Corruption rates.}
Table~\ref{tab:ablation_corruption} reports effect of varying the noise $\rho_N$ and mask rate $\rho_M$. The default configuration ($\rho_N\!=\!0.10$, $\rho_M\!=\!0.02$) is near-optimal. Removing either corruption modality degrades performance: noise-only (no mask) drops most on MMBench and TextVQA, while mask-only (no noise) shows weaker performance on SQA-IMG and MMStar. The combination of both Gaussian noising and masking consistently yields the best results, confirming that the two corruption modes provide complementary supervisory signals---continuous perturbation and discrete information removal---that together produce stronger latent denoising objective.

\begin{table}[t]
\centering
\caption{\textbf{Corruption rate ablation.} Default: $\rho_N\!=\!0.10$, $\rho_M\!=\!0.02$. All scores normalized to full latent denoising = 100.}
\label{tab:ablation_corruption}
\resizebox{0.72\columnwidth}{!}{%
\scriptsize
\setlength{\tabcolsep}{3pt}
\begin{tabular}{l|cccc}
\toprule
& high noise & high mask & no mask & no noise \\
& ($\rho_N\!=\!0.20$) & ($\rho_M\!=\!0.05$) & ($\rho_M\!=\!0$) & ($\rho_N\!=\!0$) \\
\midrule
GQA & 98.8 & 99.2 & 98.4 & 99.5 \\
POPE & 99.3 & 99.7 & 98.9 & 99.5 \\
MME & 97.7 & 97.4 & 97.5 & 99.5 \\
ChartQA & 92.9 & 94.6 & 95.1 & 98.4 \\
MMBench & 98.2 & 99.1 & 97.1 & 99.5 \\
TextVQA & 98.5 & 97.9 & 96.4 & 99.4 \\
MMStar & 99.2 & 93.1 & 94.7 & 95.0 \\
OCRBench & 97.5 & 97.5 & 97.2 & 98.5 \\
MMMU & 98.3 & 99.2 & 93.2 & 98.0 \\
SQA-IMG & 98.1 & 97.8 & 97.8 & 97.7 \\
\midrule
RWQA & 94.0 & 95.1 & 96.5 & 95.1 \\
QBench & 98.2 & 96.4 & 95.9 & 99.5 \\
VizWiz & 95.6 & 96.4 & 98.6 & 96.3 \\
NatBench-G & 89.1 & 89.1 & 90.4 & 98.1 \\
\bottomrule
\end{tabular}%
}
\end{table}

\begin{figure}[t]
    \centering
    \includegraphics[width=0.88\columnwidth]{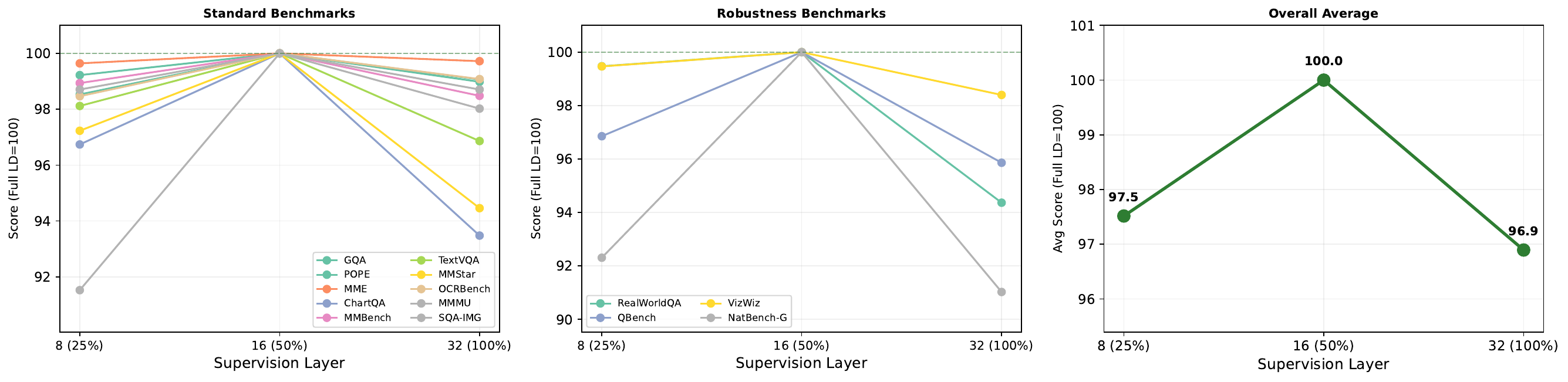}
    \caption{\textbf{Supervision layer ablation} (Full LD = 100). Mid-layer supervision (layer~16, 50\% depth) achieves the best performance across both standard and robustness benchmarks.}
    \Description{Bar chart comparing supervision at 25, 50, and 100 percent of language model depth, with mid-layer supervision performing best on standard and robustness benchmarks.}
    \label{fig:ablation_layer}
\end{figure}

\noindent \textbf{Supervision layer.}
\looseness=-1 Figure~\ref{fig:ablation_layer} shows the effect of varying the supervision layer across layers 8 (25\% depth), 16 (50\%), and 32 (100\%). Layer~16 (the default, corresponding to the mid-point of the 32-layer Vicuna backbone) achieves the best performance across nearly all benchmarks, consistent with prior observations that visual information is strongest in the middle layers of LLMs and progressively diluted at greater depth~\cite{viral,basic,fastv}. Both early and late supervision still improve over the baseline, indicating that the method is not brittle to layer choice, though mid-layer supervision provides the best balance between preserving visual signal and allowing sufficient downstream processing.

\noindent \textbf{Noise conditioning.}
Removing the noise-level embedding ($\tau$-embed) degrades performance across all benchmarks, with the largest drops on OCRBench ($-1.4$) and robustness metrics ($-1.5$ to $-2.2$), confirming that providing the model with an explicit signal of corruption magnitude is beneficial. Increasing the number of conditioning bins from 8 to 16 yields no improvement, suggesting that coarse discretization is sufficient.

\begin{figure}[t]
    \centering
    \includegraphics[width=0.75\columnwidth]{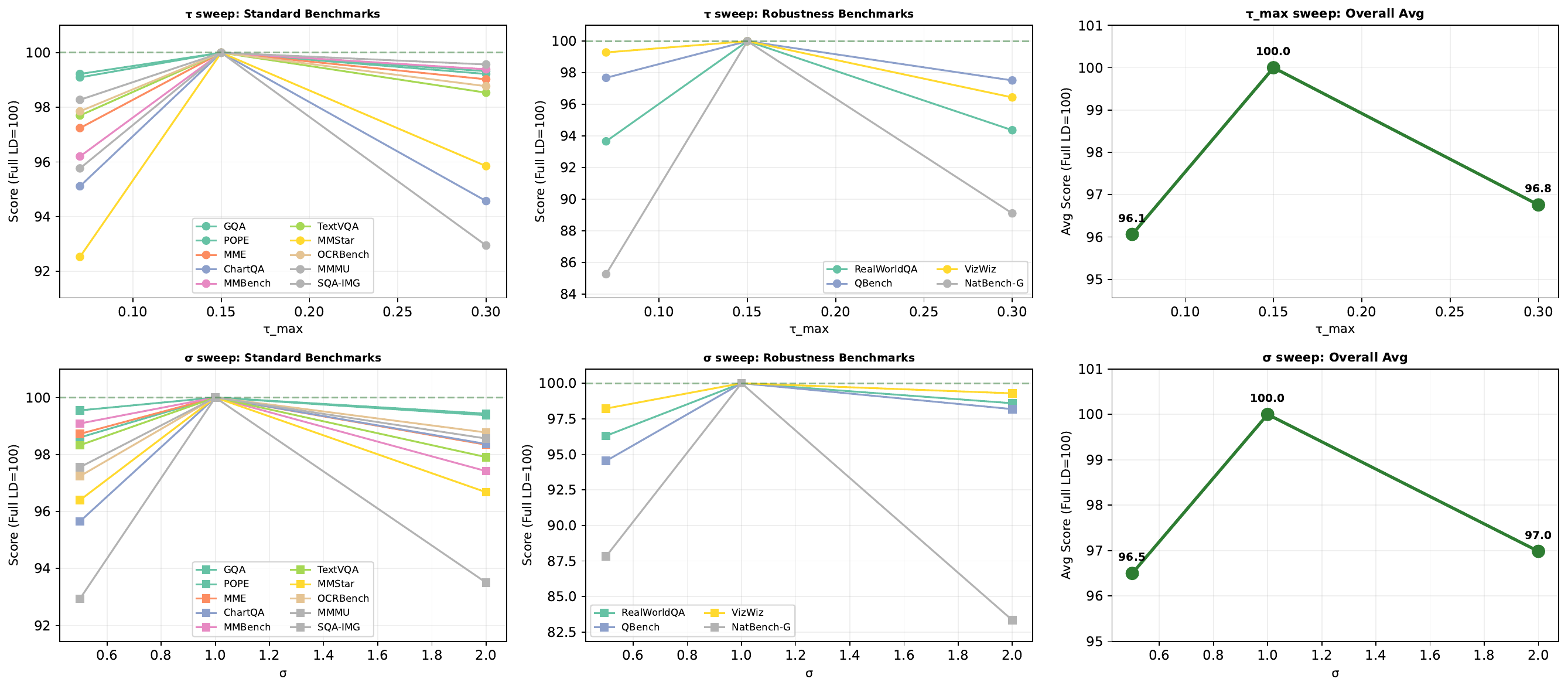}
    \caption{\textbf{Gaussian parameter sweeps} (Full LD = 100). Top: $\tau_\mathrm{max}$ sweep. Bottom: $\sigma$ sweep. Both show inverted-U patterns with default values near-optimal.}
    \Description{Line charts sweeping the Gaussian corruption parameters tau max and sigma, both showing inverted U patterns with default values near optimal.}
    \label{fig:ablation_gaussian}
\end{figure}

\noindent \textbf{Gaussian noise parameters.}
\looseness=-1 Figure~\ref{fig:ablation_gaussian} shows sweeps over $\tau_\mathrm{max}$ (interpolation strength) and $\sigma$ (noise variance). Both parameters exhibit an inverted-U pattern: the default values ($\tau_\mathrm{max}\!=\!0.15$, $\sigma\!=\!1.0$) are near-optimal, with both insufficient and excessive noise being detrimental. Too little noise ($\tau_\mathrm{max}\!=\!0.07$) provides too weak a denoising signal for the model to learn meaningful recovery, while too much ($\tau_\mathrm{max}\!=\!0.30$) overwhelms the recovery task. Notably, the method tolerates a $4\times$ range in both $\sigma$ and $\tau_\mathrm{max}$ without catastrophic failure.

\section{Conclusion}
\label{sec:concl}

\looseness=-1 We introduced a training-time latent denoising framework for LMMs that applies saliency-guided corruption to projected visual tokens and recovers aligned teacher features from intermediate hidden states, with no inference overhead. Across three architectures, 18 benchmarks, and a systematic corruption protocol, the method consistently improves visual understanding, robustness, and internal feature quality. A natural next step is composing latent denoising with complementary visual supervision paradigms~\cite{basic,viral,ross,dsvlm} at larger data and model scale.

\begin{acks}
This work is supported by the DEVCOM Army Research Lab (ARL) under grant W911NF-242-0194. Distribution Statement A: Approved for public release. Distribution is unlimited.
\end{acks}

\begingroup
\footnotesize
\def\bibfont{\footnotesize}
\setlength{\itemsep}{0pt plus 0.5pt}
\setlength{\parsep}{0pt}
\setlength{\parskip}{0pt}
\bibliographystyle{ACM-Reference-Format}
\bibliography{main}
\endgroup

\clearpage
\appendix
\section{Extended Related Works}
\label{sec:appendix_related}

\subsection{Large Multimodal Models}

The rapid progress of large language models has directly driven the development of large multimodal models (LMMs), with a dominant paradigm centered on augmenting a pretrained LLM with a visual backbone and a lightweight cross-modal connector. Representative early systems such as Flamingo, BLIP-2, InstructBLIP, MiniGPT-4, mPLUG-Owl, and LLaMA-Adapter V2 established the now-common recipe of combining a strong vision encoder with a large language model through connectors such as MLP projectors or Q-Former-style modules, followed by multimodal adaptation and instruction tuning~\cite{flamingo,blip-2,instructblip,minigpt-4,mplug-owl,llama-adapter-v2}. This design pattern has since become widespread across open-source LMM families, including LLaVA, Qwen-VL, and InternVL, with variations in visual encoders, connector design, data curation, and multi-stage training strategies~\cite{llava,llava-improved,llava-next,internvl,internvl-1.5,internvl-2.5,qwen-vl,qwen2-vl,qwen2.5-vl,qwen3-vl}. In practice, these models typically build on pretrained vision backbones such as ViT-based encoders, CLIP, and SigLIP-style vision-language encoders, while adapting the vision--language interface and language backbone through staged multimodal training~\cite{vit,clip,siglip,siglip-2}. In parallel, another line of work explores models trained more natively in a multimodal fashion through large-scale joint pretraining across modalities, including proprietary systems such as Gemini and GPT-4o and open-source or open-weight models such as Kosmos-2, Molmo, and Transfusion~\cite{gemini,gpt-4o,kosmos-2,molmo-2,transfusion}. Our work is situated in the open-source instruction-tuned LMM setting, where improving the quality of internal visual representations during multimodal training remains a central challenge.

\subsection{Visual Supervision in LMMs}

A growing body of work has sought to improve the visual capabilities of LMMs by introducing more direct supervision over their visual representations. One line of work strengthens the visual backbone itself, for example by improving locality or aligning CLIP-like features to stronger vision-centric representations before they are consumed by downstream LMMs~\cite{localityalignment,kernelclip}. While effective, these methods primarily refine the visual encoder or its feature space outside the full multimodal instruction-tuning process. A second line of work injects supervision more directly into the multimodal stack by refining projector outputs or internal visual tokens using shallow-layer embeddings, patch-level alignment, or guidance from one or more vision foundation models~\cite{basic,finegrained,viral,vaco}. These approaches more explicitly target visual alignment within the LMM, but are largely formulated as clean feature alignment or token refinement objectives. A third line introduces reconstructive or generative supervision through latent reconstruction, masked image modeling, or diffusion-based objectives~\cite{ross,laver,dsvlm}. Compared with these methods, our work is motivated by a different perspective: rather than treating visual supervision purely as backbone refinement, direct feature alignment, or generative reconstruction, we cast it as \emph{generalized latent denoising} for internal visual representations during multimodal instruction tuning.

\subsection{Denoising for Visual Representation Learning}

Corruption-and-recovery objectives have long served as a foundational principle for visual representation learning. Early denoising autoencoders learned useful representations by reconstructing clean signals from corrupted inputs, while later masked image modeling methods showed that masking-based recovery can scale effectively to high-capacity vision transformers~\cite{dae,mae}. In parallel, modern diffusion-based generative models have made Gaussian denoising the dominant paradigm for high-fidelity image synthesis, and recent work has further extended this perspective to semantically meaningful latent spaces through diffusion autoencoders and related latent generative frameworks~\cite{ddpm,ldm}. Most closely related to our motivation, recent tokenizer work argues that combining masking and Gaussian corruption in latent space yields stronger visual tokenizers for downstream generation, suggesting denoising as a general design principle rather than a generation-specific trick~\cite{detok}. Our work is inspired by this broader view, but departs from prior denoising-based visual learning in a key way: we bring generalized latent denoising into multimodal instruction tuning, using it to directly supervise internal visual features inside large multimodal models for visual understanding.

\subsection{Understanding--Generation Bridge}

Recent work increasingly suggests that visual understanding and visual generation should not be viewed as entirely separate objectives, but rather as two perspectives on learning strong visual representations, as reflected in unified mixed-modal models, joint multimodal pretraining frameworks, and unified visual tokenizer designs~\cite{tokenflow,transfusion}. In generation, a long line of work has shown that the quality of latent spaces and visual tokenizers critically determines downstream synthesis quality, from discrete visual tokenization and masked token modeling to modern latent diffusion and representation autoencoders~\cite{vqvae,vqgan,ldm,rae,scale-rae}. More recent multimodal pretraining and unified multimodal modeling work further strengthens this connection by showing that semantically meaningful visual representations can simultaneously benefit understanding and generation within shared training pipelines~\cite{transfusion}. In parallel, recent tokenizer work shows that the quality of visual latents is a primary determinant of downstream generation quality, and that latent denoising in particular can provide an especially effective mechanism for improving visual tokenizers~\cite{detok,tokenflow,textok,rae,scale-rae}. Our work is motivated by this broader bridge: while principles from visual understanding are increasingly used to improve visual generation, the reverse direction remains underexplored in large multimodal models.

\subsection{Corruption Evaluation in LMMs}

Robustness to corrupted visual inputs has been studied extensively in vision, where the common-corruptions paradigm established by ImageNet-C provides a standardized, non-adversarial benchmark across noise, blur, weather, and digital distortions~\cite{hendrycks2019benchmarking}. More recently, this perspective has begun to extend to multimodal settings. In particular, recent work introduces large-scale visual robustness evaluation for visual question answering under realistic corruptions~\cite{vqarobustness}, while other studies benchmark the real-world robustness of large multimodal models under broader image degradation pipelines and user-capture distortions~\cite{rbench}. In parallel, a distinct line of work studies \emph{adversarial} robustness in vision-language models, focusing on worst-case perturbations, prompt-based defenses, or adversarially motivated adaptation strategies~\cite{attackvlm,zeroshotadvvlm}. Our work is aligned with the common-corruptions tradition rather than the adversarial setting: we adapt the ImageNet-C-style corruption paradigm into a deterministic, family-structured evaluation protocol for LMMs, enabling systematic assessment under non-adversarial corruptions across a broad range of multimodal benchmarks.

\section{Additional Method Details}
\label{sec:appendix_method}

We provide supplementary details for the latent denoising framework introduced in
Sec.~\ref{sec:method}, focusing on saliency extraction, asymmetric sampling,
noise conditioning, model-specific implementation choices, and the full training
algorithm in Algorithm~\ref{alg:latent_denoising}.

\subsection{Saliency Extraction and Asymmetric Sampling}
\label{sec:appendix_saliency}

Our corruption policy is guided by patch saliency derived from the frozen visual
encoder. For ViT-style encoders with a classification token, we use the
class-to-patch attention from the same encoder layer used to extract the visual
features:
\begin{equation}
s_i=\frac{1}{H}\sum_{h=1}^{H}\mathbf{A}^{(\ell)}_{h,\mathrm{cls}\rightarrow i},
\label{eq:saliency_cls}
\end{equation}
where $\mathbf{A}^{(\ell)}\in\mathbb{R}^{H\times(S+1)\times(S+1)}$ denotes the multi-head
attention matrix at encoder layer $\ell$. For encoders without a class token, we
instead use the average attention received by patch $i$:
\begin{equation}
s_i=\frac{1}{HS}\sum_{h=1}^{H}\sum_{j=1}^{S}\mathbf{A}^{(\ell)}_{h,j\rightarrow i},
\label{eq:saliency_recv}
\end{equation}
with $\mathbf{A}^{(\ell)}\in\mathbb{R}^{H\times S\times S}$. In architectures where
class-token attention is not exposed in the same form, we use an encoder-native
saliency surrogate computed directly from the visual features.

Given saliency scores $\{s_i\}_{i=1}^{S}$, we sample two disjoint patch sets:
a noised set $\mathcal{N}$ of size
$K_N=\lfloor \rho_N S \rfloor$ and a masked set $\mathcal{M}$ of size
$K_M=\lfloor \rho_M S \rfloor$. High-saliency patches are preferentially
assigned to $\mathcal{N}$ using
\begin{equation}
\mathbf{p}_{\mathcal{N}}(i)=\frac{\exp(s_i/\tau_s)}{\sum_{k=1}^{S}\exp(s_k/\tau_s)},
\label{eq:p_high}
\end{equation}
whereas lower-saliency patches are preferentially assigned to $\mathcal{M}$ via
the complementary distribution
\begin{equation}
\mathbf{p}_{\mathcal{M}}(i)=\frac{\exp(-s_i/\tau_s)}{\sum_{k\notin \mathcal{N}}\exp(-s_k/\tau_s)},
\qquad i\notin \mathcal{N}.
\label{eq:p_low}
\end{equation}
Sampling is performed without replacement, first for $\mathcal{N}$ and then for
$\mathcal{M}$ from the remaining patch set. This asymmetric strategy softens
corruption on visually important regions while reserving hard masking for less
salient content.

\subsection{Noise Conditioning and Auxiliary Loss Details}
\label{sec:appendix_conditioning}

For continuously noised tokens, we further add a learned noise-level embedding.
Each sampled corruption level $\tau_i$ is discretized into one of $B$ bins,
\begin{equation}
b_i=\left\lfloor \frac{\tau_i}{\tau_{\max}}B \right\rfloor,
\qquad b_i \in \{0,\dots,B-1\},
\label{eq:tau_bins}
\end{equation}
and a learned embedding
$\mathbf{e}_{\tau}(b_i)\in\mathbb{R}^{d_h}$ is added to the corrupted token
representation for $i\in\mathcal{N}$. Masked and clean tokens do not receive
this conditioning. In practice, this embedding provides the model with an
explicit signal of corruption magnitude while preserving the same inference path
as the base LMM.

All auxiliary losses are computed only on corrupted tokens. The reconstruction
term in Eq.~\eqref{eq:lrec} uses L2-normalized predictions and targets, making
it equivalent (up to a constant factor) to cosine distance rather than raw MSE.
The relational term in Eq.~\eqref{eq:lrel} uses the teacher similarity
distribution as the reference and minimizes a
$\mathrm{KL}(\text{teacher}\,\|\,\text{student})$ divergence. The contrastive
term in Eq.~\eqref{eq:lcon} is an asymmetric student-to-teacher InfoNCE
objective with negatives drawn from other corrupted patches within the same
image.

\subsection{Model-Specific Implementation Notes}
\label{sec:appendix_model_specific}

Our framework is instantiated across both LLaVA-style and Qwen-style LMMs, with
minor architecture-specific differences.

\paragraph{Teacher feature space.}
In LLaVA-style models, the visual features used for projection, the teacher
targets, and the saliency signal are all extracted from the same penultimate
layer of the frozen visual encoder, following the standard design used in
LLaVA-like models. The decoder therefore maps from the LMM hidden space back to
the pre-projector visual feature space. In Qwen-style models, by contrast, we do
not use the raw final-layer ViT features before the visual merger. In practice,
we found these features to be highly degenerate and nearly collinear across
patches, making them unsuitable as reconstruction targets. Instead, we use the
post-merger visual features directly as teacher targets, which yields a much more
informative representation space for latent denoising. This choice is consistent
with prior observations that deeper transformer features can become
over-globalized or collapse in diversity, and that visual signal often weakens
substantially at greater depth \cite{fastv,revit}.

\paragraph{Saliency source.}
In CLIP- and SigLIP-based LLaVA variants, saliency is derived from visual
attention patterns. In Qwen-style models, we instead use a feature-space
saliency surrogate computed directly from the post-merger visual representations,
which is better aligned with the visual space actually consumed by the language
model.

\paragraph{Visual token layout.}
LLaVA-style models use a fixed set of projected image tokens inserted into the
language sequence, whereas Qwen-style models naturally accommodate variable
numbers of image tokens per image. Despite these implementation differences, the
conceptual method remains unchanged: saliency-guided corruption is applied in the
LMM visual-token space, and auxiliary supervision is imposed on internal visual
hidden states.

\begin{algorithm}[t]
\caption{Latent Denoising for Internal Visual Features}
\label{alg:latent_denoising}
\begin{algorithmic}[1]
\Require image $\mathbf{X}$, frozen vision encoder $E_{\boldsymbol{\phi}}$, frozen teacher $F_{\boldsymbol{\zeta}}$, projector $P_{\boldsymbol{\psi}}$, LMM $L_{\boldsymbol{\theta}}$, decoder $D_{\boldsymbol{\chi}}$
\Require corruption rates $(\rho_N,\rho_M)$, corruption parameters $(\sigma,\tau_{\max})$, temperatures $(\tau_s,\tau_r,\tau_c)$, loss weights $(\lambda_{\mathrm{rec}},\lambda_{\mathrm{rel}},\lambda_{\mathrm{con}})$
\State Extract visual features $\mathbf{V}=E_{\boldsymbol{\phi}}(\mathbf{X})$ and detached teacher targets $\mathbf{Y}=F_{\boldsymbol{\zeta}}(\mathbf{X})$
\State Compute patch saliency scores $\{s_i\}_{i=1}^{S}$ from the frozen visual encoder
\State Sample a high-saliency noised set $\mathcal{N}$ and a low-saliency masked set $\mathcal{M}$
\State Project visual features into the LMM embedding space: $\mathbf{z}_i=P_{\boldsymbol{\psi}}(\mathbf{v}_i)$
\For{$i=1,\dots,S$}
    \If{$i\in\mathcal{M}$}
        \State $\tilde{\mathbf{z}}_i \gets \mathbf{e}_{\mathrm{mask}}$
    \ElsIf{$i\in\mathcal{N}$}
        \State Sample $\tau_i \!\sim\! \mathrm{Unif}(0,\tau_{\max})$ and $\boldsymbol{\epsilon}_i \!\sim\! \mathcal{N}(\mathbf{0},\sigma^2\mathbf{I})$
        \State $\tilde{\mathbf{z}}_i \gets (1-\tau_i)\mathbf{z}_i+\tau_i\boldsymbol{\epsilon}_i$
        \State Add noise-level embedding $\mathbf{e}_{\tau}(b_i)$
    \Else
        \State $\tilde{\mathbf{z}}_i \gets \mathbf{z}_i$
    \EndIf
\EndFor
\State Run the LMM on the corrupted visual tokens and text tokens
\State Extract visual hidden states from the selected supervision layer $\ell_s$
\State Decode corrupted-token states: $\hat{\mathbf{y}}_i = D_{\boldsymbol{\chi}}(\mathbf{h}_i^{(\ell_s)})$ for $i\in\mathcal{C}=\mathcal{N}\cup\mathcal{M}$
\State Compute $\mathcal{L}_{\mathrm{rec}}$, $\mathcal{L}_{\mathrm{rel}}$, and $\mathcal{L}_{\mathrm{con}}$ on the corrupted-token set
\State Combine with the language modeling loss to obtain the final objective
\end{algorithmic}
\end{algorithm}

\section{Additional Corruption Results}
\label{sec:appendix_corruption}

The main paper reports per-corruption-type results at severity~5 for LLaVA+CLIP (Table~\ref{tab:corruption_clip}) and Qwen-2.5-VL (Table~\ref{tab:corruption_qwen}). Here we provide the remaining tables to complete the evaluation across all architectures and severity levels.

\noindent \textbf{LLaVA+CLIP, severity~3} (Table~\ref{tab:appendix_clip_s3}). Latent denoising improves on all 11 benchmarks across all four corruption types at the moderate severity level, with the same directional patterns as severity~5.

\begin{table}
\centering
\caption{\textbf{Severity~3, LLaVA+CLIP.} B = Baseline, LD = Latent Denoising.}
\label{tab:appendix_clip_s3}
\resizebox{\columnwidth}{!}{%
\scriptsize
\setlength{\tabcolsep}{2pt}
\begin{tabular}{l|ccc|ccc|ccc|ccc}
\toprule
& \multicolumn{3}{c|}{\textbf{Noise}} & \multicolumn{3}{c|}{\textbf{Blur}} & \multicolumn{3}{c|}{\textbf{Weather}} & \multicolumn{3}{c}{\textbf{Digital}} \\
& B & LD & $\Delta$ & B & LD & $\Delta$ & B & LD & $\Delta$ & B & LD & $\Delta$ \\
\midrule
VQAv2 & 74.3 & \textbf{75.4} & \cmark{+1.1} & 71.2 & \textbf{72.3} & \cmark{+1.1} & 73.7 & \textbf{75.0} & \cmark{+1.3} & 75.8 & \textbf{76.9} & \cmark{+1.1} \\
GQA & 60.2 & \textbf{62.4} & \cmark{+2.2} & 59.1 & \textbf{60.7} & \cmark{+1.6} & 60.1 & \textbf{62.3} & \cmark{+2.2} & 61.4 & \textbf{63.5} & \cmark{+2.1} \\
POPE & 84.2 & \textbf{85.7} & \cmark{+1.5} & 81.8 & \textbf{83.4} & \cmark{+1.6} & 84.0 & \textbf{85.8} & \cmark{+1.8} & 85.5 & \textbf{87.1} & \cmark{+1.6} \\
MME & 1767 & \textbf{1783} & \cmark{+17} & 1811 & \textbf{1825} & \cmark{+13} & 1798 & \textbf{1809} & \cmark{+12} & 1815 & \textbf{1826} & \cmark{+12} \\
MMB & 61.5 & \textbf{63.1} & \cmark{+1.6} & 58.3 & \textbf{59.5} & \cmark{+1.2} & 61.0 & \textbf{63.0} & \cmark{+2.0} & 63.1 & \textbf{64.1} & \cmark{+1.0} \\
MMStar & 33.5 & \textbf{37.3} & \cmark{+3.8} & 32.1 & \textbf{36.0} & \cmark{+3.9} & 33.1 & \textbf{37.8} & \cmark{+4.7} & 33.9 & \textbf{36.7} & \cmark{+2.8} \\
MMMU & 35.1 & \textbf{37.6} & \cmark{+2.5} & 35.1 & \textbf{36.4} & \cmark{+1.3} & 35.6 & \textbf{37.0} & \cmark{+1.4} & 35.0 & \textbf{36.7} & \cmark{+1.7} \\
SQA & 68.9 & \textbf{69.3} & \cmark{+0.4} & 69.3 & \textbf{70.1} & \cmark{+0.8} & 67.7 & \textbf{68.5} & \cmark{+0.8} & 69.3 & \textbf{69.8} & \cmark{+0.5} \\
TxtVQA & 42.0 & \textbf{43.1} & \cmark{+1.1} & 31.4 & \textbf{32.4} & \cmark{+1.0} & 41.0 & \textbf{42.1} & \cmark{+1.1} & 44.0 & \textbf{45.2} & \cmark{+1.2} \\
OCR & 25.0 & \textbf{26.1} & \cmark{+1.1} & 17.0 & \textbf{18.4} & \cmark{+1.4} & 29.0 & \textbf{30.0} & \cmark{+1.0} & 26.9 & \textbf{28.2} & \cmark{+1.3} \\
Chart & 16.0 & \textbf{17.0} & \cmark{+1.0} & 13.2 & \textbf{14.0} & \cmark{+0.8} & 16.8 & \textbf{17.4} & \cmark{+0.6} & 16.5 & \textbf{17.3} & \cmark{+0.8} \\
\bottomrule
\end{tabular}%
}
\end{table}

\noindent \textbf{LLaVA+SigLIP, severity~3} (Table~\ref{tab:appendix_siglip_s3}) \textbf{and severity~5} (Table~\ref{tab:appendix_siglip_s5}). The SigLIP variant shows the same consistent improvement pattern, confirming cross-encoder generalization of the corruption robustness benefit.

\begin{table}
\centering
\caption{\textbf{Severity~3, LLaVA+SigLIP.}}
\label{tab:appendix_siglip_s3}
\resizebox{\columnwidth}{!}{%
\scriptsize
\setlength{\tabcolsep}{2pt}
\begin{tabular}{l|ccc|ccc|ccc|ccc}
\toprule
& \multicolumn{3}{c|}{\textbf{Noise}} & \multicolumn{3}{c|}{\textbf{Blur}} & \multicolumn{3}{c|}{\textbf{Weather}} & \multicolumn{3}{c}{\textbf{Digital}} \\
& B & LD & $\Delta$ & B & LD & $\Delta$ & B & LD & $\Delta$ & B & LD & $\Delta$ \\
\midrule
VQAv2 & 73.3 & \textbf{74.5} & \cmark{+1.2} & 69.8 & \textbf{71.2} & \cmark{+1.4} & 72.6 & \textbf{73.9} & \cmark{+1.3} & 74.5 & \textbf{76.2} & \cmark{+1.7} \\
GQA & 59.5 & \textbf{61.5} & \cmark{+2.0} & 57.8 & \textbf{60.4} & \cmark{+2.6} & 60.0 & \textbf{62.0} & \cmark{+2.0} & 60.9 & \textbf{62.8} & \cmark{+1.9} \\
POPE & 83.8 & \textbf{85.3} & \cmark{+1.5} & 80.7 & \textbf{82.4} & \cmark{+1.7} & 84.3 & \textbf{85.6} & \cmark{+1.3} & 85.1 & \textbf{86.4} & \cmark{+1.3} \\
MME & 1677 & \textbf{1694} & \cmark{+17} & 1646 & \textbf{1666} & \cmark{+21} & 1699 & \textbf{1715} & \cmark{+16} & 1681 & \textbf{1701} & \cmark{+20} \\
MMB & 60.9 & \textbf{63.2} & \cmark{+2.3} & 55.3 & \textbf{58.1} & \cmark{+2.8} & 62.0 & \textbf{63.7} & \cmark{+1.7} & 63.4 & \textbf{66.1} & \cmark{+2.7} \\
MMStar & 34.0 & \textbf{36.1} & \cmark{+2.1} & 32.5 & \textbf{33.1} & \cmark{+0.6} & 35.2 & \textbf{37.4} & \cmark{+2.2} & 34.9 & \textbf{36.7} & \cmark{+1.8} \\
MMMU & 33.8 & \textbf{35.9} & \cmark{+2.1} & 33.0 & \textbf{34.8} & \cmark{+1.8} & 34.1 & \textbf{36.0} & \cmark{+1.9} & 34.2 & \textbf{35.9} & \cmark{+1.7} \\
SQA & 69.6 & \textbf{70.0} & \cmark{+0.4} & 69.0 & \textbf{70.0} & \cmark{+1.0} & 68.4 & \textbf{69.4} & \cmark{+1.0} & 70.5 & \textbf{71.0} & \cmark{+0.5} \\
TxtVQA & 43.6 & \textbf{45.7} & \cmark{+2.1} & 32.3 & \textbf{34.2} & \cmark{+1.9} & 43.1 & \textbf{44.7} & \cmark{+1.6} & 46.9 & \textbf{48.7} & \cmark{+1.8} \\
OCR & 26.4 & \textbf{27.9} & \cmark{+1.5} & 19.1 & \textbf{20.4} & \cmark{+1.3} & 29.3 & \textbf{30.8} & \cmark{+1.5} & 28.5 & \textbf{30.3} & \cmark{+1.8} \\
Chart & 14.8 & \textbf{16.2} & \cmark{+1.4} & 13.0 & \textbf{14.5} & \cmark{+1.5} & 15.2 & \textbf{16.6} & \cmark{+1.4} & 14.9 & \textbf{16.0} & \cmark{+1.1} \\
\bottomrule
\end{tabular}%
}
\end{table}

\begin{table}
\centering
\caption{\textbf{Severity~5, LLaVA+SigLIP.}}
\label{tab:appendix_siglip_s5}
\resizebox{\columnwidth}{!}{%
\scriptsize
\setlength{\tabcolsep}{2pt}
\begin{tabular}{l|ccc|ccc|ccc|ccc}
\toprule
& \multicolumn{3}{c|}{\textbf{Noise}} & \multicolumn{3}{c|}{\textbf{Blur}} & \multicolumn{3}{c|}{\textbf{Weather}} & \multicolumn{3}{c}{\textbf{Digital}} \\
& B & LD & $\Delta$ & B & LD & $\Delta$ & B & LD & $\Delta$ & B & LD & $\Delta$ \\
\midrule
VQAv2 & 66.9 & \textbf{69.1} & \cmark{+2.2} & 65.6 & \textbf{68.5} & \cmark{+2.9} & 69.9 & \textbf{72.4} & \cmark{+2.5} & 69.5 & \textbf{71.7} & \cmark{+2.2} \\
GQA & 55.7 & \textbf{58.5} & \cmark{+2.8} & 56.0 & \textbf{58.9} & \cmark{+2.9} & 57.9 & \textbf{60.8} & \cmark{+2.9} & 57.5 & \textbf{60.1} & \cmark{+2.6} \\
POPE & 78.4 & \textbf{81.5} & \cmark{+3.1} & 77.6 & \textbf{80.4} & \cmark{+2.8} & 82.4 & \textbf{85.0} & \cmark{+2.6} & 79.9 & \textbf{82.6} & \cmark{+2.7} \\
MME & 1597 & \textbf{1625} & \cmark{+27} & 1523 & \textbf{1562} & \cmark{+38} & 1623 & \textbf{1640} & \cmark{+18} & 1631 & \textbf{1659} & \cmark{+28} \\
MMB & 52.2 & \textbf{55.3} & \cmark{+3.1} & 50.3 & \textbf{53.1} & \cmark{+2.8} & 54.9 & \textbf{58.1} & \cmark{+3.2} & 55.5 & \textbf{58.3} & \cmark{+2.8} \\
MMStar & 30.0 & \textbf{32.8} & \cmark{+2.8} & 31.1 & \textbf{32.8} & \cmark{+1.7} & 32.0 & \textbf{35.3} & \cmark{+3.3} & 32.2 & \textbf{34.5} & \cmark{+2.3} \\
MMMU & 32.6 & \textbf{35.3} & \cmark{+2.7} & 31.8 & \textbf{33.4} & \cmark{+1.6} & 32.5 & \textbf{35.4} & \cmark{+2.9} & 32.6 & \textbf{35.1} & \cmark{+2.5} \\
SQA & 67.7 & \textbf{69.7} & \cmark{+2.0} & 67.0 & \textbf{69.7} & \cmark{+2.7} & 67.6 & \textbf{69.5} & \cmark{+1.9} & 68.4 & \textbf{69.8} & \cmark{+1.4} \\
TxtVQA & 35.9 & \textbf{38.6} & \cmark{+2.7} & 24.4 & \textbf{27.3} & \cmark{+2.9} & 40.6 & \textbf{43.1} & \cmark{+2.5} & 37.5 & \textbf{40.0} & \cmark{+2.5} \\
OCR & 18.6 & \textbf{21.7} & \cmark{+3.1} & 13.3 & \textbf{15.9} & \cmark{+2.6} & 26.6 & \textbf{29.6} & \cmark{+3.0} & 21.5 & \textbf{24.4} & \cmark{+2.9} \\
Chart & 12.6 & \textbf{15.1} & \cmark{+2.5} & 12.2 & \textbf{14.7} & \cmark{+2.5} & 14.1 & \textbf{16.8} & \cmark{+2.7} & 13.2 & \textbf{15.4} & \cmark{+2.2} \\
\bottomrule
\end{tabular}%
}
\end{table}

\noindent \textbf{Qwen-2.5-VL, severity~3} (Table~\ref{tab:appendix_qwen_s3}). GQA shows the largest gains (+5.8 to +6.5), consistent with the severity~5 pattern in the main paper.

\begin{table}
\centering
\caption{\textbf{Severity~3, Qwen-2.5-VL.}}
\label{tab:appendix_qwen_s3}
\resizebox{\columnwidth}{!}{%
\scriptsize
\setlength{\tabcolsep}{2pt}
\begin{tabular}{l|ccc|ccc|ccc|ccc}
\toprule
& \multicolumn{3}{c|}{\textbf{Noise}} & \multicolumn{3}{c|}{\textbf{Blur}} & \multicolumn{3}{c|}{\textbf{Weather}} & \multicolumn{3}{c}{\textbf{Digital}} \\
& B & LD & $\Delta$ & B & LD & $\Delta$ & B & LD & $\Delta$ & B & LD & $\Delta$ \\
\midrule
VQAv2 & 73.5 & \textbf{74.7} & \cmark{+1.2} & 71.5 & \textbf{73.0} & \cmark{+1.5} & 76.2 & \textbf{77.6} & \cmark{+1.4} & 77.8 & \textbf{79.0} & \cmark{+1.2} \\
GQA & 55.8 & \textbf{62.0} & \cmark{+6.2} & 54.5 & \textbf{61.0} & \cmark{+6.5} & 58.2 & \textbf{64.2} & \cmark{+6.0} & 59.0 & \textbf{64.8} & \cmark{+5.8} \\
POPE & 81.2 & \textbf{83.0} & \cmark{+1.8} & 79.5 & \textbf{81.8} & \cmark{+2.3} & 84.3 & \textbf{86.2} & \cmark{+1.9} & 83.2 & \textbf{85.4} & \cmark{+2.2} \\
MME & 2218 & \textbf{2233} & \cmark{+15} & 2121 & \textbf{2148} & \cmark{+28} & 2311 & \textbf{2326} & \cmark{+15} & 2285 & \textbf{2309} & \cmark{+23} \\
MMB & 72.9 & \textbf{74.7} & \cmark{+1.8} & 68.3 & \textbf{70.6} & \cmark{+2.3} & 75.8 & \textbf{77.3} & \cmark{+1.5} & 77.2 & \textbf{79.5} & \cmark{+2.3} \\
MMStar & 54.5 & \textbf{56.1} & \cmark{+1.6} & 51.0 & \textbf{52.4} & \cmark{+1.4} & 56.8 & \textbf{58.7} & \cmark{+1.9} & 56.5 & \textbf{57.6} & \cmark{+1.1} \\
MMMU & 48.7 & \textbf{51.2} & \cmark{+2.5} & 46.2 & \textbf{48.8} & \cmark{+2.6} & 49.5 & \textbf{51.8} & \cmark{+2.3} & 49.0 & \textbf{51.4} & \cmark{+2.4} \\
SQA & 83.3 & \textbf{84.4} & \cmark{+1.1} & 81.2 & \textbf{82.8} & \cmark{+1.6} & 84.8 & \textbf{85.5} & \cmark{+0.7} & 85.6 & \textbf{86.8} & \cmark{+1.2} \\
TxtVQA & 62.0 & \textbf{63.2} & \cmark{+1.2} & 48.5 & \textbf{50.1} & \cmark{+1.6} & 72.4 & \textbf{74.5} & \cmark{+2.1} & 72.8 & \textbf{74.2} & \cmark{+1.4} \\
OCR & 60.1 & \textbf{61.5} & \cmark{+1.4} & 42.8 & \textbf{44.6} & \cmark{+1.8} & 70.2 & \textbf{72.4} & \cmark{+2.2} & 62.5 & \textbf{64.0} & \cmark{+1.5} \\
Chart & 64.0 & \textbf{67.6} & \cmark{+3.6} & 48.5 & \textbf{50.3} & \cmark{+1.8} & 72.1 & \textbf{75.3} & \cmark{+3.2} & 66.8 & \textbf{69.0} & \cmark{+2.2} \\
\bottomrule
\end{tabular}%
}
\end{table}

\section{Full Ablation Tables}
\label{sec:appendix_ablations}

We provide the complete ablation results across all benchmarks, complementing the plots and selected results in the main paper. All ablations use LoRA fine-tuning on the CLIP+Vicuna configuration.

\noindent \textbf{Loss components} (Table~\ref{tab:app_loss}). The full latent denoising objective outperforms the baseline across nearly all benchmarks. Removing any single loss or switching to random token selection consistently reduces performance. ``Rand.'' uses all three losses but with random (non-saliency) token selection.

\begin{table}
\centering
\caption{\textbf{Loss component ablation.} $r\!=\!\mathcal{L}_\mathrm{rec}$, $e\!=\!\mathcal{L}_\mathrm{rel}$, $c\!=\!\mathcal{L}_\mathrm{con}$. BL = no auxiliary losses. Best \textbf{bold}.}
\label{tab:app_loss}
\scriptsize
\setlength{\tabcolsep}{3pt}
\begin{tabular}{l|cccccc}
\toprule
& BL & LD & $r$ & $r\!+\!e$ & $r\!+\!c$ & Rand. \\
\midrule
VQAv2 & 77.3 & \textbf{78.1} & 77.1 & 77.1 & 77.0 & 77.0 \\
GQA & 63.3 & \textbf{64.2} & 63.3 & 62.9 & 63.3 & 63.0 \\
POPE & 87.5 & \textbf{88.4} & 87.4 & 87.9 & 87.5 & 87.9 \\
MME & 1747 & \textbf{1786} & 1696 & 1747 & 1722 & 1668 \\
MMB & 64.9 & \textbf{65.8} & 64.8 & 65.3 & 64.3 & 65.0 \\
TxtVQA & 47.1 & \textbf{47.8} & 46.7 & 47.0 & 46.7 & 46.9 \\
MMStar & 34.5 & \textbf{36.1} & 34.9 & 34.6 & 33.7 & 33.5 \\
OCR & 31.9 & \textbf{32.6} & 31.0 & 30.9 & 30.6 & 30.5 \\
MMMU & 34.6 & \textbf{35.4} & 34.0 & 32.9 & 33.8 & 33.1 \\
SQA & \textbf{69.8} & 69.5 & 71.0 & 69.2 & 70.7 & 69.4 \\
Chart & 17.2 & \textbf{18.4} & 18.0 & 18.0 & 18.0 & 17.2 \\
\midrule
RWQA & 56.2 & \textbf{56.8} & 56.5 & 55.0 & 56.1 & 55.4 \\
QBench & 59.2 & \textbf{60.4} & 60.1 & 59.4 & 58.8 & 57.5 \\
VizWiz & 55.6 & \textbf{56.2} & 50.2 & 51.8 & 54.1 & 55.7 \\
NatB-Q & 41.4 & \textbf{41.8} & 40.1 & 38.1 & 39.9 & 38.5 \\
NatB-I & 46.4 & \textbf{46.9} & 45.3 & 44.0 & 45.7 & 44.3 \\
NatB-G & 14.9 & \textbf{15.6} & 15.0 & 13.7 & 13.9 & 13.6 \\
\bottomrule
\end{tabular}
\end{table}

\noindent \textbf{Supervision layer} (Table~\ref{tab:app_layer}). Layer~16 (50\% depth, default) achieves the best performance. Both early (layer~8) and late (layer~32) supervision still improve over the baseline.

\begin{table}
\centering
\caption{\textbf{Student layer ablation.} Default: layer~16 (50\%).}
\label{tab:app_layer}
\scriptsize
\setlength{\tabcolsep}{4pt}
\begin{tabular}{l|ccc}
\toprule
& Layer 8 (25\%) & Layer 16 (50\%) & Layer 32 (100\%) \\
\midrule
GQA & 63.7 & \textbf{64.2} & 63.6 \\
POPE & 87.1 & \textbf{88.4} & 87.5 \\
MME & 1779 & \textbf{1786} & 1781 \\
Chart & 17.8 & \textbf{18.4} & 17.2 \\
MMB & 65.1 & \textbf{65.8} & 64.8 \\
TextVQA & 46.9 & \textbf{47.8} & 46.3 \\
MMStar & 35.1 & \textbf{36.1} & 34.1 \\
OCR & 32.1 & \textbf{32.6} & 32.3 \\
MMMU & 32.4 & \textbf{35.4} & 34.7 \\
SQA & 68.6 & \textbf{69.5} & 68.6 \\
\midrule
RWQA & 56.5 & \textbf{56.8} & 53.6 \\
QBench & 58.5 & \textbf{60.4} & 57.9 \\
VizWiz & 55.9 & \textbf{56.2} & 55.3 \\
NatB-G & 14.4 & \textbf{15.6} & 14.2 \\
\bottomrule
\end{tabular}
\end{table}

\noindent \textbf{Noise conditioning} (Table~\ref{tab:app_cond}). Removing the noise-level embedding degrades performance across all benchmarks. Increasing bins from 8 to 16 provides no benefit.

\begin{table}
\centering
\caption{\textbf{Noise conditioning ablation.} Default: 8 bins with $\tau$-embed.}
\label{tab:app_cond}
\scriptsize
\setlength{\tabcolsep}{4pt}
\begin{tabular}{l|ccc}
\toprule
& No Embed & Default (8) & 16 Bins \\
\midrule
GQA & 63.5 & \textbf{64.2} & 63.5 \\
POPE & 87.7 & \textbf{88.4} & 87.7 \\
MME & 1741 & \textbf{1786} & 1723 \\
Chart & 17.6 & \textbf{18.4} & 17.2 \\
MMB & 64.8 & \textbf{65.8} & 64.6 \\
TextVQA & 46.8 & \textbf{47.8} & 46.8 \\
MMStar & 34.7 & \textbf{36.1} & 34.8 \\
OCR & 31.2 & \textbf{32.6} & 32.2 \\
MMMU & 34.3 & \textbf{35.4} & 33.8 \\
SQA & 68.6 & \textbf{69.5} & 67.7 \\
\midrule
RWQA & 55.3 & \textbf{56.8} & 55.1 \\
QBench & 58.3 & \textbf{60.4} & 58.1 \\
VizWiz & 54.3 & \textbf{56.2} & 55.4 \\
NatB-G & 13.4 & \textbf{15.6} & 13.6 \\
\bottomrule
\end{tabular}
\end{table}

\noindent \textbf{Gaussian noise parameters} (Table~\ref{tab:app_gauss}). Both $\tau_\mathrm{max}$ and $\sigma$ exhibit inverted-U patterns with default values near-optimal. The method tolerates a $4\times$ range in both parameters without catastrophic degradation.

\begin{table}
\centering
\caption{\textbf{Gaussian parameter ablation.} Default: $\tau_\mathrm{max}\!=\!0.15$, $\sigma\!=\!1.0$.}
\label{tab:app_gauss}
\scriptsize
\setlength{\tabcolsep}{4pt}
\begin{tabular}{l|cccc}
\toprule
& $\tau_\mathrm{max}\!=\!.07$ & $\tau_\mathrm{max}\!=\!.30$ & $\sigma\!=\!0.5$ & $\sigma\!=\!2.0$ \\
\midrule
GQA & 63.7 & 63.7 & 63.3 & 63.8 \\
POPE & 87.6 & 87.8 & 88.0 & 87.9 \\
MME & 1736 & 1768 & 1763 & 1756 \\
Chart & 17.5 & 17.4 & 17.6 & 18.1 \\
MMB & 63.3 & 65.4 & 65.2 & 64.1 \\
TextVQA & 46.7 & 47.1 & 47.0 & 46.8 \\
MMStar & 33.4 & 34.6 & 34.8 & 34.9 \\
OCR & 31.9 & 32.2 & 31.7 & 32.2 \\
MMMU & 33.9 & 32.9 & 32.9 & 33.1 \\
SQA & 68.3 & 69.2 & 67.8 & 68.5 \\
\midrule
RWQA & 53.2 & 53.6 & 54.7 & 56.0 \\
QBench & 59.0 & 58.9 & 57.1 & 59.3 \\
VizWiz & 55.8 & 54.2 & 55.2 & 55.8 \\
NatB-G & 13.3 & 13.9 & 13.7 & 13.0 \\
\bottomrule
\end{tabular}
\end{table}

\section{Training Details and Hyperparameters}
\label{sec:appendix_training}

\noindent \textbf{LLaVA with CLIP and SigLIP.}
We follow the standard LLaVA-1.5 two-stage recipe~\cite{llava-improved}. Stage~1 (projector pretraining): MLP projector trained on LLaVA-558K with vision encoder and LLM frozen, learning rate $1\!\times\!10^{-3}$, cosine schedule, warmup ratio 0.03, effective batch size 128 (per-device 4 $\times$ gradient accumulation 4 $\times$ 8 GPUs), 1 epoch, DeepSpeed ZeRO-2. Latent denoising is disabled in stage~1. Stage~2 (instruction tuning): projector and full LLM unfrozen on LLaVA-mix-665K, learning rate $2\!\times\!10^{-5}$, cosine schedule, same batch size, 1 epoch, DeepSpeed ZeRO-2. The CLIP variant uses ViT-L/14@336 (576 patches, 1024-dim); the SigLIP variant uses SO400M/14@384 (729 patches, 1152-dim). Both use Vicuna-7B-v1.5 as the language backbone. Matched baselines are trained with identical settings but latent denoising disabled.

\noindent \textbf{Qwen-2.5-VL.}
We post-tune the publicly released Qwen-2.5-VL-7B-Instruct checkpoint on LLaVA-mix-665K for 1 epoch with latent denoising enabled. Learning rate $2\!\times\!10^{-5}$, cosine schedule, effective batch size 128, DeepSpeed ZeRO-2. The visual merger output features serve as both teacher targets and saliency source, as described in Sec.~\ref{sec:appendix_model_specific}.

\noindent \textbf{Latent denoising parameters.}
Table~\ref{tab:app_ld_params} lists all latent denoising hyperparameters, shared across architectures unless noted. The decoder is a two-layer MLP with GELU activation, mapping from $d_h$ to $d_t$.

\begin{table}
\centering
\caption{\textbf{Latent denoising hyperparameters.}}
\label{tab:app_ld_params}
\scriptsize
\setlength{\tabcolsep}{4pt}
\begin{tabular}{llc}
\toprule
\textbf{Parameter} & \textbf{Description} & \textbf{Value} \\
\midrule
$\rho_N$ & Noise rate & 0.10 \\
$\rho_M$ & Mask rate & 0.02 \\
$\sigma$ & Noise variance & 1.0 \\
$\tau_\mathrm{max}$ & Max interpolation strength & 0.15 \\
$\tau_s$ & Saliency temperature & 0.07 \\
$\tau_r$ & Relational temperature & 0.10 \\
$\tau_c$ & Contrastive temperature & 0.07 \\
$B$ & Noise conditioning bins & 8 \\
$\lambda_\mathrm{rec}$ & Reconstruction loss weight & 0.10 \\
$\lambda_\mathrm{rel}$ & Relational loss weight & 0.025 \\
$\lambda_\mathrm{con}$ & Contrastive loss weight & 0.025 \\
\midrule
\multicolumn{3}{l}{\textit{WHD Schedule}} \\
Warmup & & 5\% \\
Hold & & 75\% \\
Decay & & 20\% \\
\midrule
Supervision layer & & 16 (50\% depth) \\
\bottomrule
\end{tabular}
\end{table}

\noindent \textbf{LoRA ablations.}
All ablation experiments use LoRA~\cite{lora} with rank $r\!=\!128$, $\alpha\!=\!256$, dropout $0.05$, applied to all linear layers in the LLM. The projector, decoder, and auxiliary embeddings (mask, noise-level) remain fully trainable. Learning rate $2\!\times\!10^{-4}$ for LoRA parameters, $2\!\times\!10^{-5}$ for the projector. All other settings match stage~2.

\noindent \textbf{Evaluation.}
All evaluations use \texttt{lmms-eval} \cite{lmms-eval} with batch size 1 and greedy decoding. Corruption evaluation follows the protocol in Sec.~\ref{sec:exp_corruption}.

\noindent \textbf{WHD schedule details.}
The warmup--hold--decay (WHD) schedule linearly increases corruption rates and auxiliary loss weights from zero during an initial warmup phase (5\% of training), holds them at target values for the majority of training (75\%), and linearly decays them back to zero in the final phase (20\%). This prevents auxiliary losses from interfering with the critical early optimization steps of the language model and ensures clean convergence at the end of training.

\noindent \textbf{Corruption evaluation protocol details.}
Within each corruption family, specific subtypes are: \emph{noise} (Gaussian, shot, impulse, speckle), \emph{blur} (defocus, glass, motion, zoom, Gaussian), \emph{weather} (snow, frost, fog, brightness, spatter), and \emph{digital} (contrast, elastic, pixelate, JPEG, saturate). For each image, a single subtype is deterministically selected using a seeded hash of the image identifier, ensuring reproducibility and one-to-one correspondence across models evaluated on the same benchmark. We evaluate at severity levels 3 (moderate) and 5 (severe) on the standard 1--5 ImageNet-C scale.

\section{Comparison with Previous Works}
\label{sec:appendix_comparison}

Table~\ref{tab:app_method_comparison} provides a qualitative comparison of our method with recent approaches that introduce visual supervision during LMM training, spanning backbone refinement~\cite{localityalignment,kernelclip}, feature alignment~\cite{basic,finegrained,viral,vaco}, and reconstructive supervision~\cite{ross,laver,dsvlm} (see Sec.~\ref{sec:related}). Direct quantitative comparison is difficult as these methods use different evaluation protocols, benchmark subsets, and training configurations.

\begin{table}
\centering
\caption{\textbf{Comparison with visual supervision methods for LMMs.} VFM = vision foundation model.}
\label{tab:app_method_comparison}
\resizebox{\columnwidth}{!}{%
\scriptsize
\setlength{\tabcolsep}{2.5pt}
\begin{tabular}{lccc}
\toprule
\textbf{Method} & \textbf{Supervision Type} & \textbf{Teacher} & \textbf{Training Overhead} \\
\midrule
LocalAlign~\cite{localityalignment} & Backbone alignment & DINO/SAM & Encoder fine-tuning \\
KernelCLIP~\cite{kernelclip} & Kernel alignment & DINOv2 & Encoder fine-tuning \\
\midrule
BASIC~\cite{basic} & Embedding alignment & Self (LLM) & Minimal \\
FineGrained~\cite{finegrained} & Patch-level alignment & External VFM & Projector head \\
ViRAL~\cite{viral} & Feature regularization & DINOv2 (frozen) & Frozen VFM forward pass \\
VaCo~\cite{vaco} & Multi-VFM coordination & Multiple VFMs & Multiple VFM forward passes \\
\midrule
ROSS~\cite{ross} & Diffusion reconstruction & Encoder & Diffusion head \\
LaVER~\cite{laver} & Masked reconstruction & EMA copy of LLM & Full model EMA \\
DS-VLM~\cite{dsvlm} & Diffusion reconstruction & Encoder & Diffusion head \\
\midrule
\textbf{Ours} & Latent denoising & Backbone encoder & Lightweight MLP decoder \\
\bottomrule
\end{tabular}%
}
\end{table}

Our method is lightweight by design: the teacher signal comes from the same frozen vision encoder already used as the LMM's backbone, requiring no additional external model at training time. The auxiliary decoder is a two-layer MLP, substantially cheaper than the diffusion heads used by ROSS and DS-VLM or the full-model EMA required by LaVER. In contrast, ViRAL and VaCo require forward passes through one or more frozen vision foundation models (e.g., DINOv2, SAM) at every training step. While an external teacher VFM could in principle be used, our framework achieves strong results using only the backbone encoder, keeping the method self-contained.

\section{Additional Visual Feature Analysis}
\label{sec:appendix_features}

\begin{figure}[t]
    \centering
    \includegraphics[width=\columnwidth]{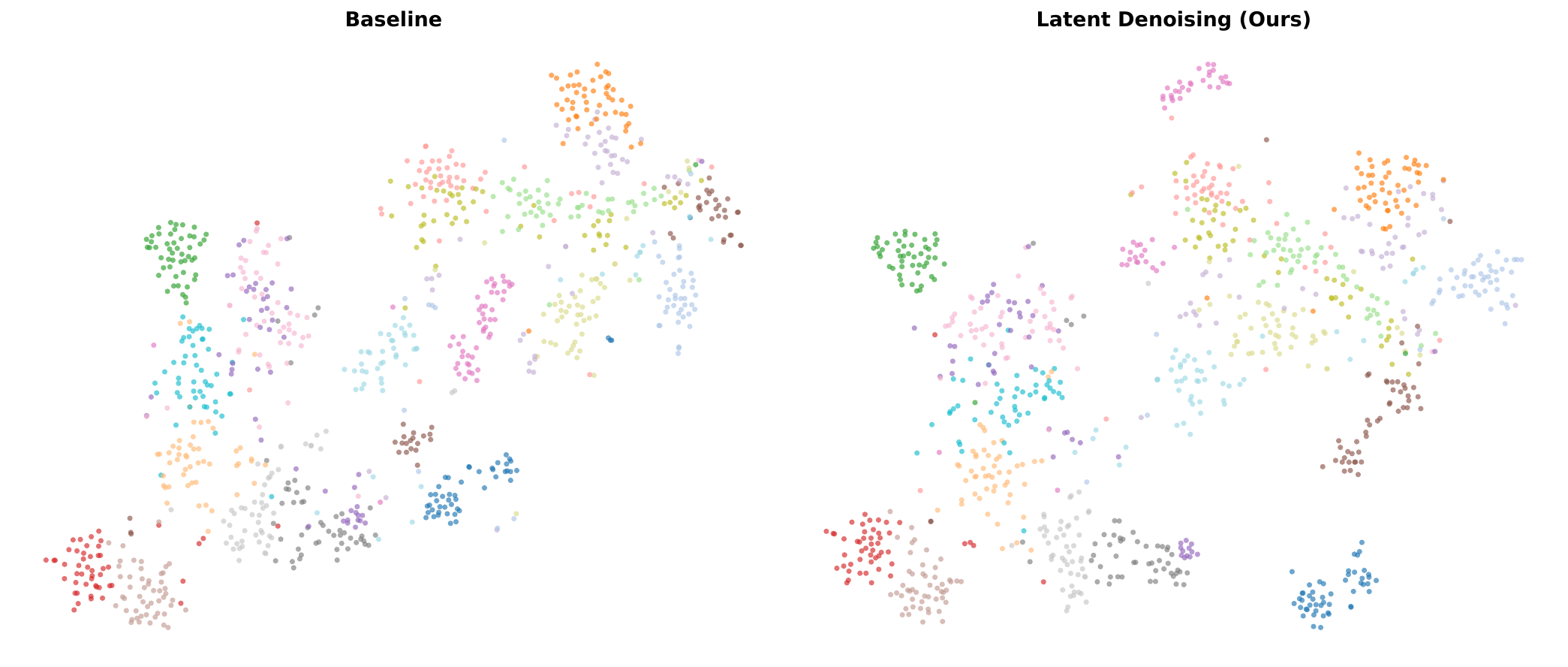}
    \caption{\textbf{t-SNE at layer 15} (supervised layer). Latent denoising produces tighter class clusters.}
    \label{fig:app_tsne_15}
\end{figure}

\begin{figure}[t]
    \centering
    \includegraphics[width=\columnwidth]{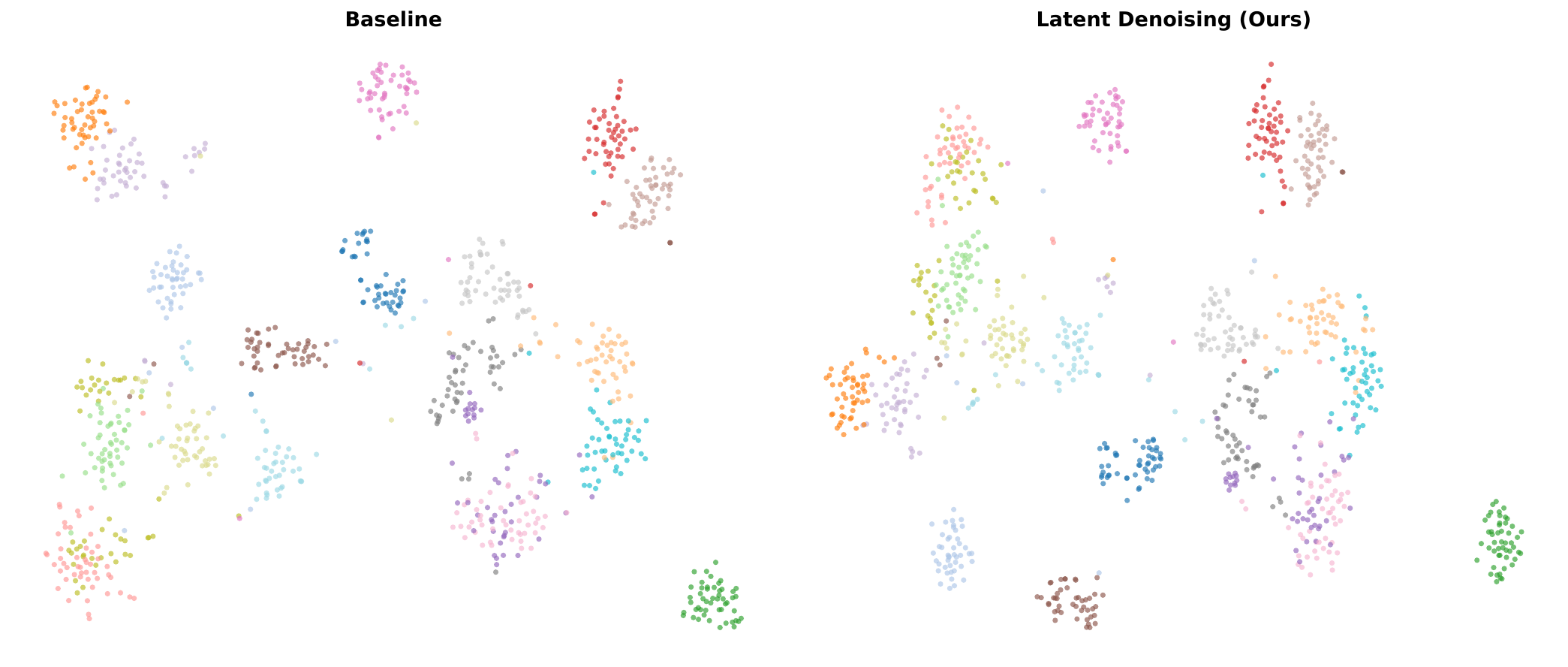}
    \caption{\textbf{t-SNE at layer 24.} Cluster structure is maintained at deeper layers.}
    \label{fig:app_tsne_24}
\end{figure}

\begin{figure}[t]
    \centering
    \includegraphics[width=\columnwidth]{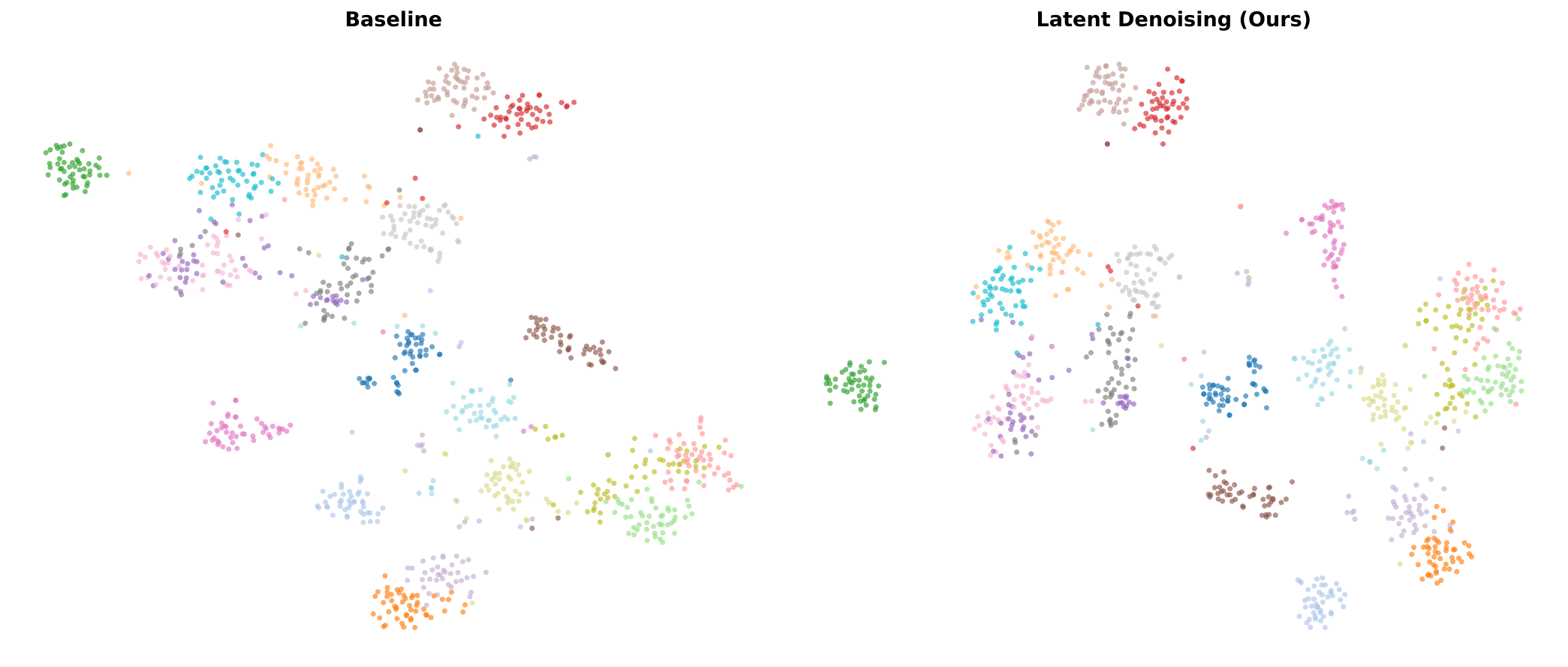}
    \caption{\textbf{t-SNE at layer 31} (final layer). Latent denoising preserves discriminative structure even at the output layer.}
    \label{fig:app_tsne_31}
\end{figure}

\noindent \textbf{t-SNE visualizations.}
Figures~\ref{fig:app_tsne_15}--\ref{fig:app_tsne_31} show t-SNE visualizations of mean-pooled visual features from 20 ImageNet classes (50 images each) at layers 15, 24, and 31. At all depths, latent denoising produces tighter, more separated class clusters, consistent with the kNN classification improvements reported in the main paper.

\begin{figure*}[p]
    \centering
    \includegraphics[width=0.85\textwidth,trim=0 18 0 18,clip]{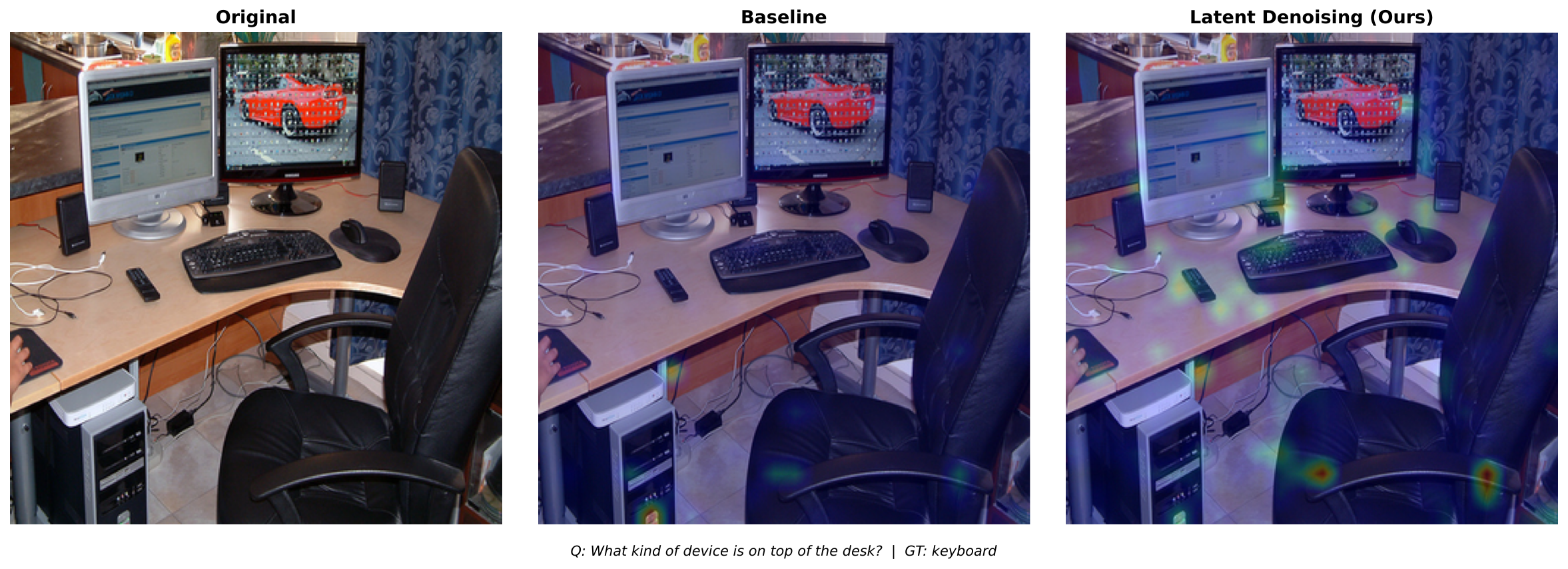}\\[-2pt]
    {\scriptsize\itshape Q: What kind of device is on top of the desk? $\mid$ GT: keyboard}\\[1pt]
    \includegraphics[width=0.85\textwidth,trim=0 18 0 18,clip]{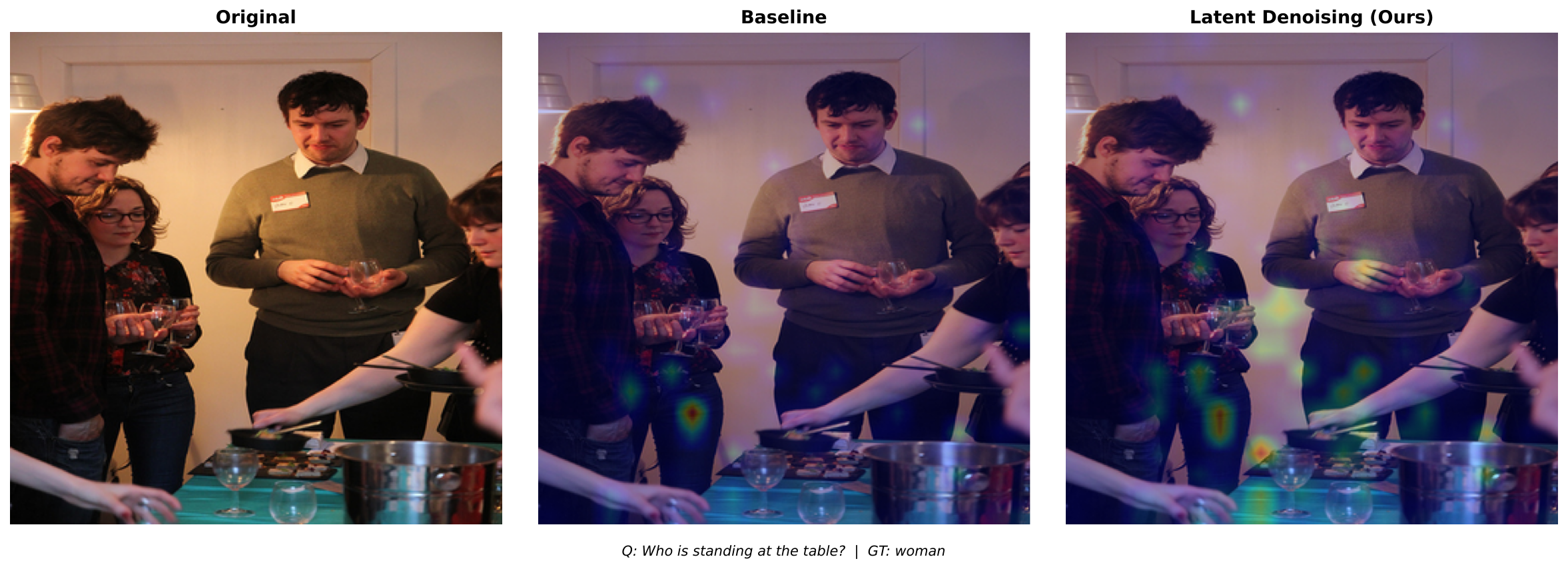}\\[-2pt]
    {\scriptsize\itshape Q: Who is standing at the table? $\mid$ GT: woman}\\[1pt]
    \includegraphics[width=0.85\textwidth,trim=0 18 0 18,clip]{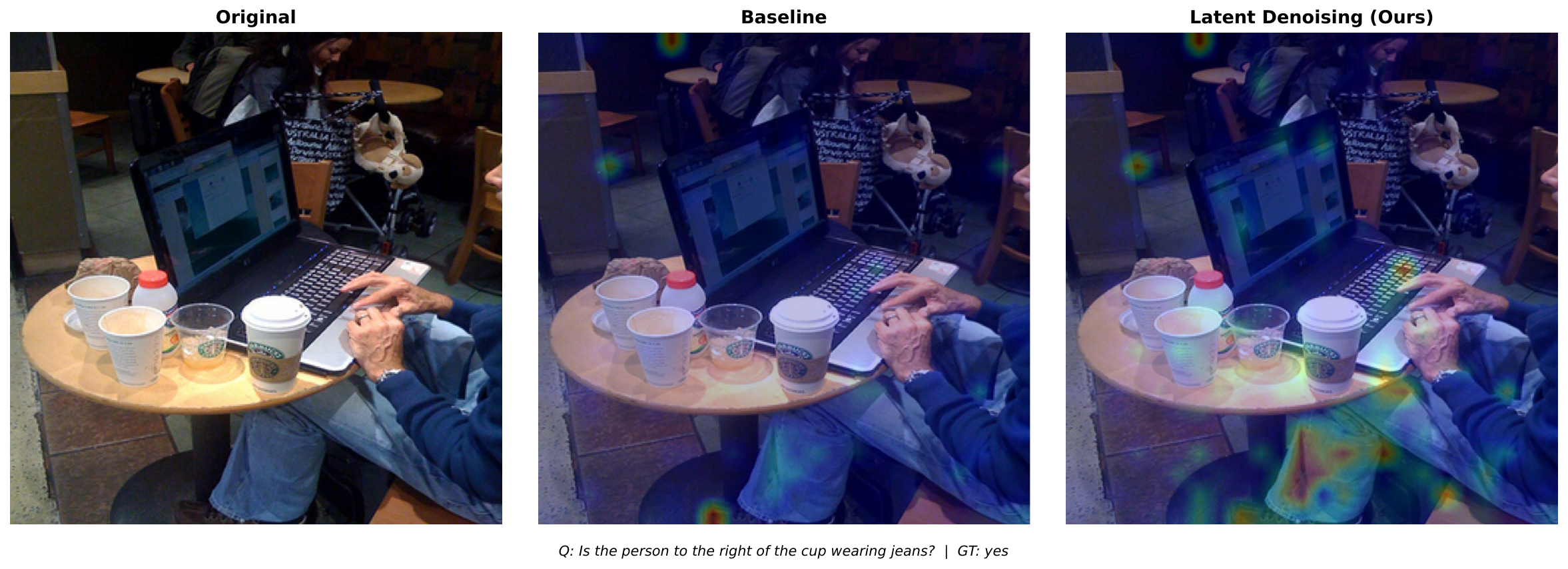}\\[-2pt]
    {\scriptsize\itshape Q: Is the person to the right of the cup wearing jeans? $\mid$ GT: yes}\\[1pt]
    \includegraphics[width=0.85\textwidth,trim=0 18 0 18,clip]{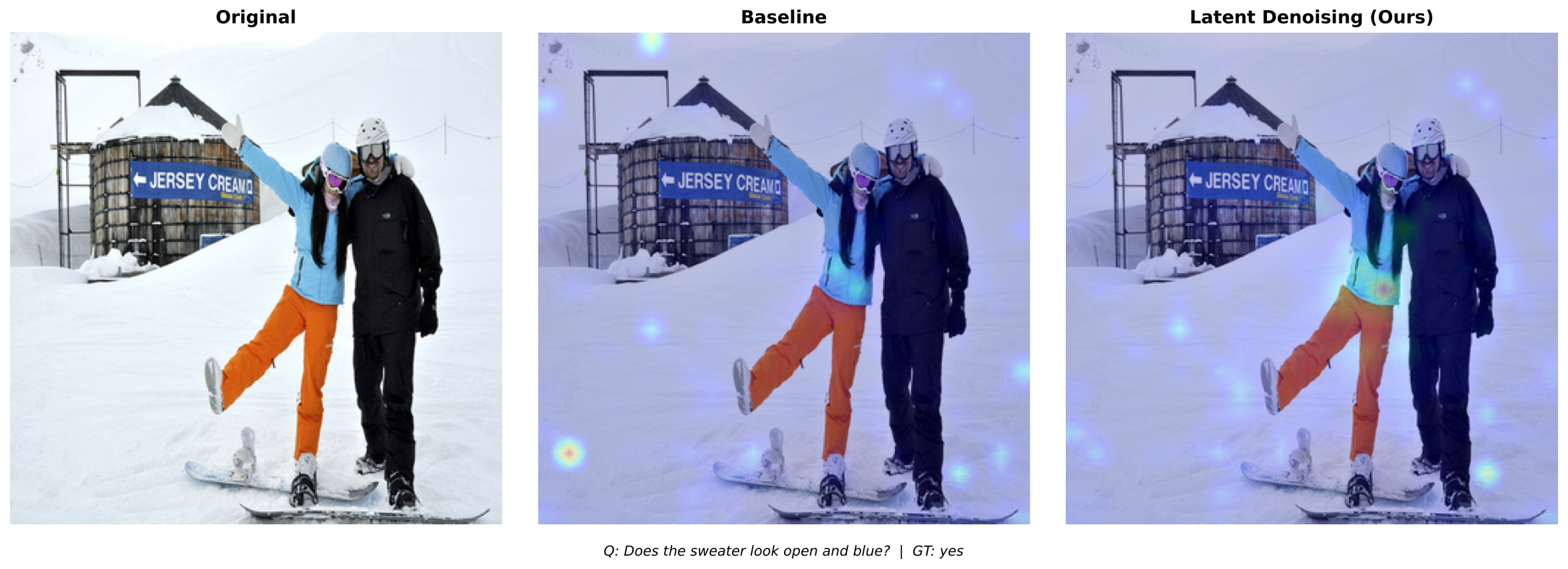}\\[-2pt]
    {\scriptsize\itshape Q: Does the sweater look open and blue? $\mid$ GT: yes}
    \caption{\textbf{Additional cross-attention heatmaps (1/3).} Layer~16, GQA. Each row: original, baseline, latent denoising.}
    \label{fig:app_attn_1}
\end{figure*}

\begin{figure*}[p]
    \centering
    \includegraphics[width=0.85\textwidth,trim=0 18 0 18,clip]{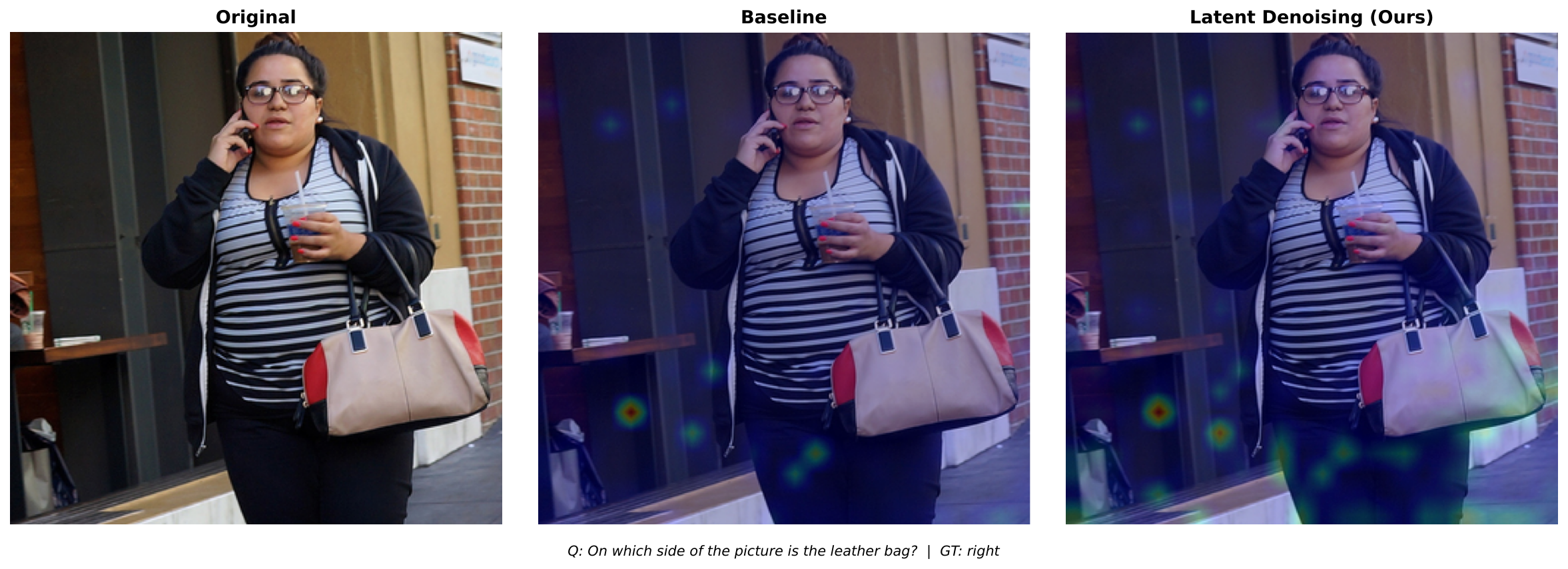}\\[-2pt]
    {\scriptsize\itshape Q: On which side of the picture is the leather bag? $\mid$ GT: right}\\[1pt]
    \includegraphics[width=0.85\textwidth,trim=0 18 0 18,clip]{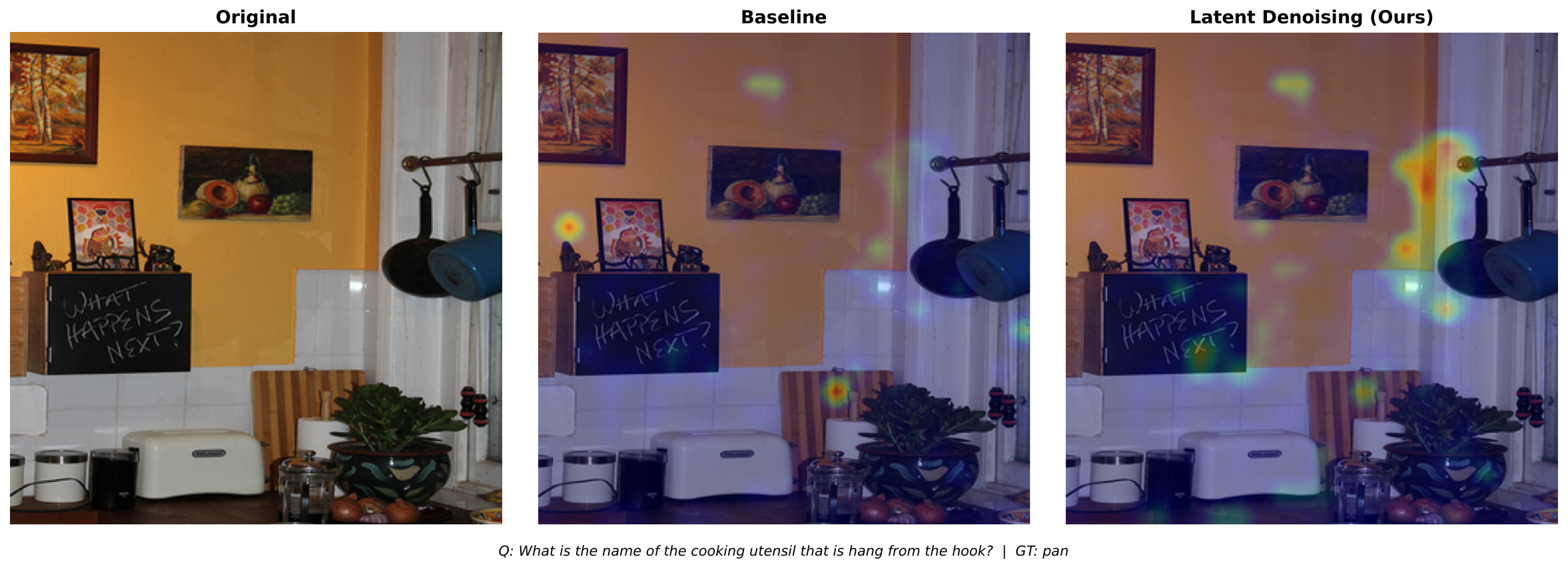}\\[-2pt]
    {\scriptsize\itshape Q: What is the cooking utensil hanging from the hook? $\mid$ GT: pan}\\[1pt]
    \includegraphics[width=0.85\textwidth,trim=0 18 0 18,clip]{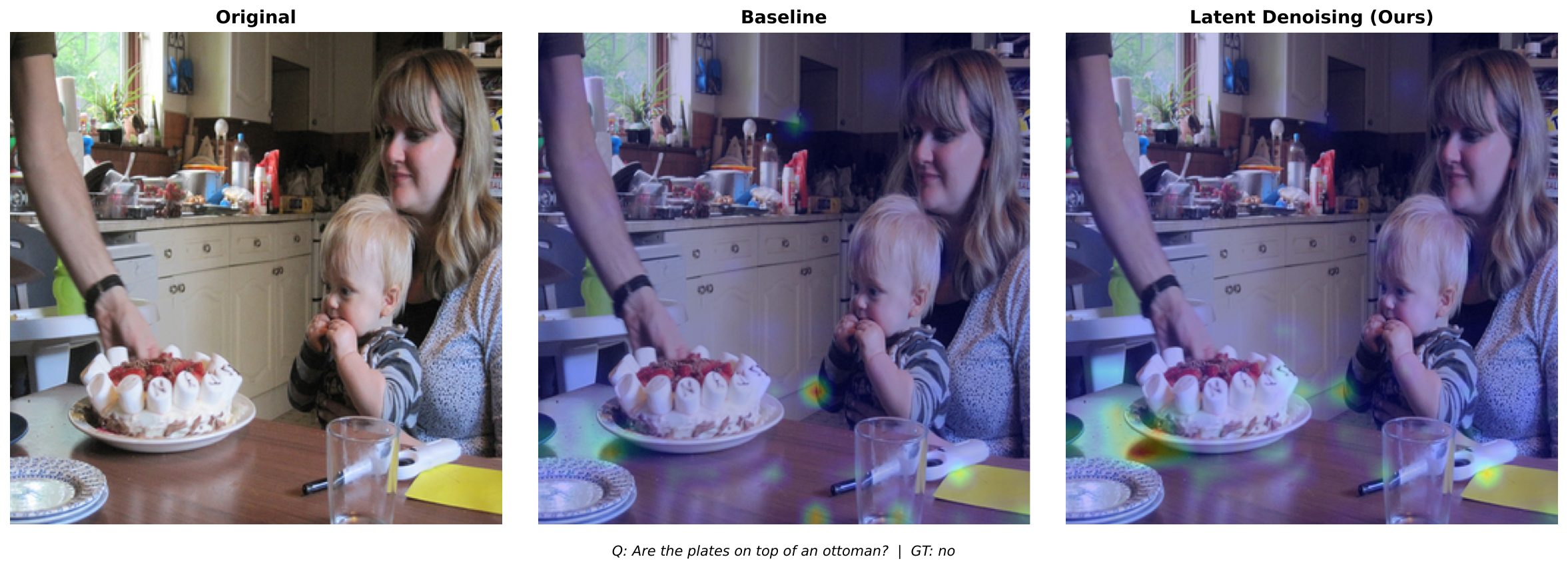}\\[-2pt]
    {\scriptsize\itshape Q: Are the plates on top of an ottoman? $\mid$ GT: no}\\[1pt]
    \includegraphics[width=0.85\textwidth,trim=0 18 0 18,clip]{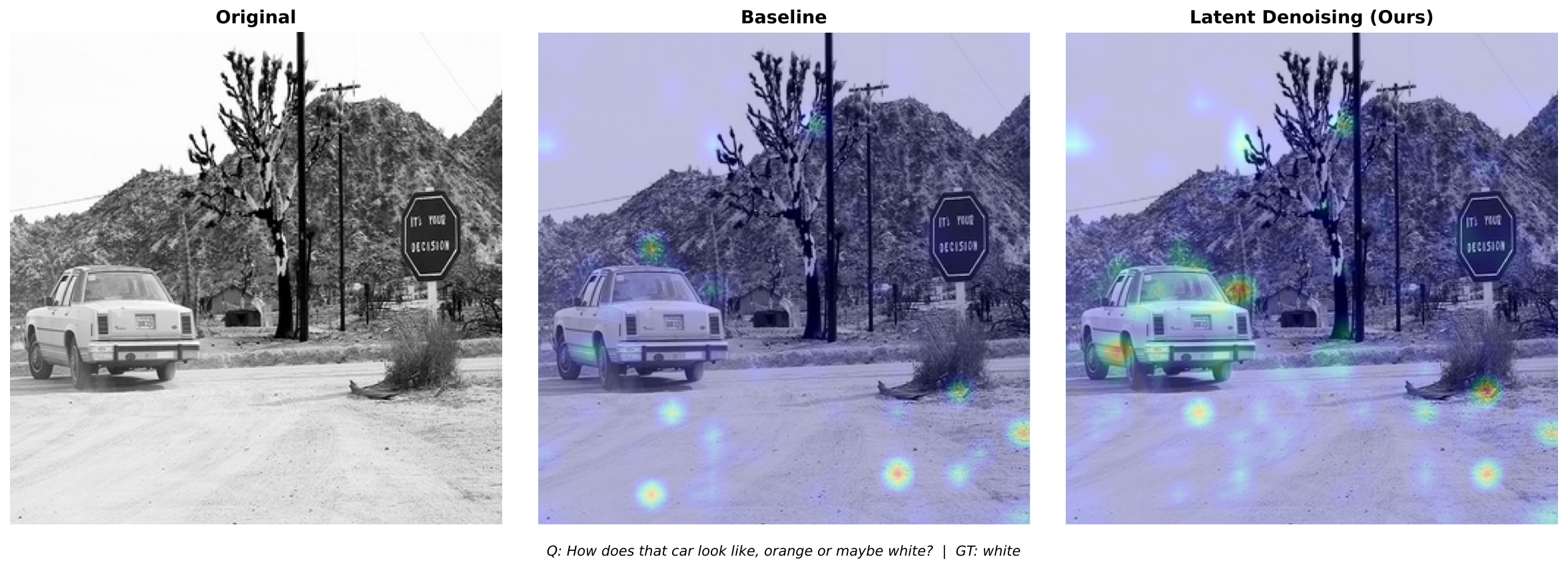}\\[-2pt]
    {\scriptsize\itshape Q: How does that car look, orange or maybe white? $\mid$ GT: white}
    \caption{\textbf{Additional cross-attention heatmaps (2/3).} Layer~16, GQA.}
    \label{fig:app_attn_2}
\end{figure*}

\begin{figure*}[p]
    \centering
    \includegraphics[width=0.85\textwidth,trim=0 18 0 18,clip]{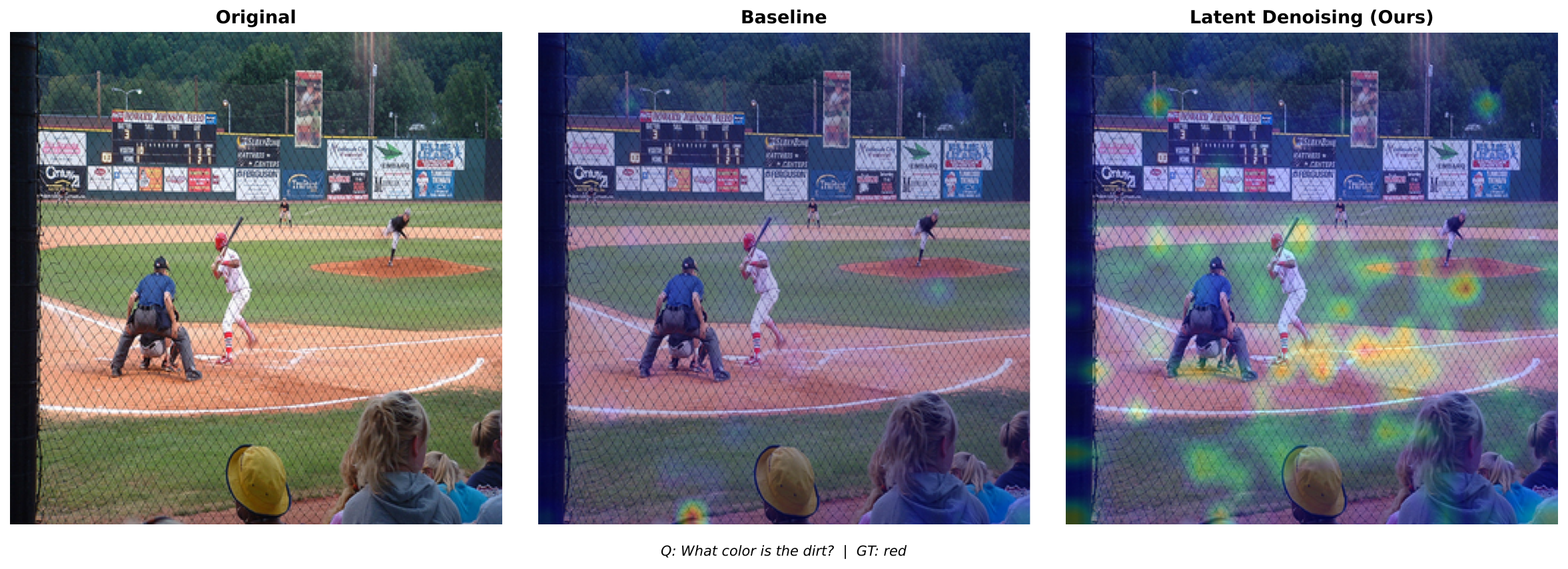}\\[-2pt]
    {\scriptsize\itshape Q: What color is the dirt? $\mid$ GT: red}\\[1pt]
    \includegraphics[width=0.85\textwidth,trim=0 18 0 18,clip]{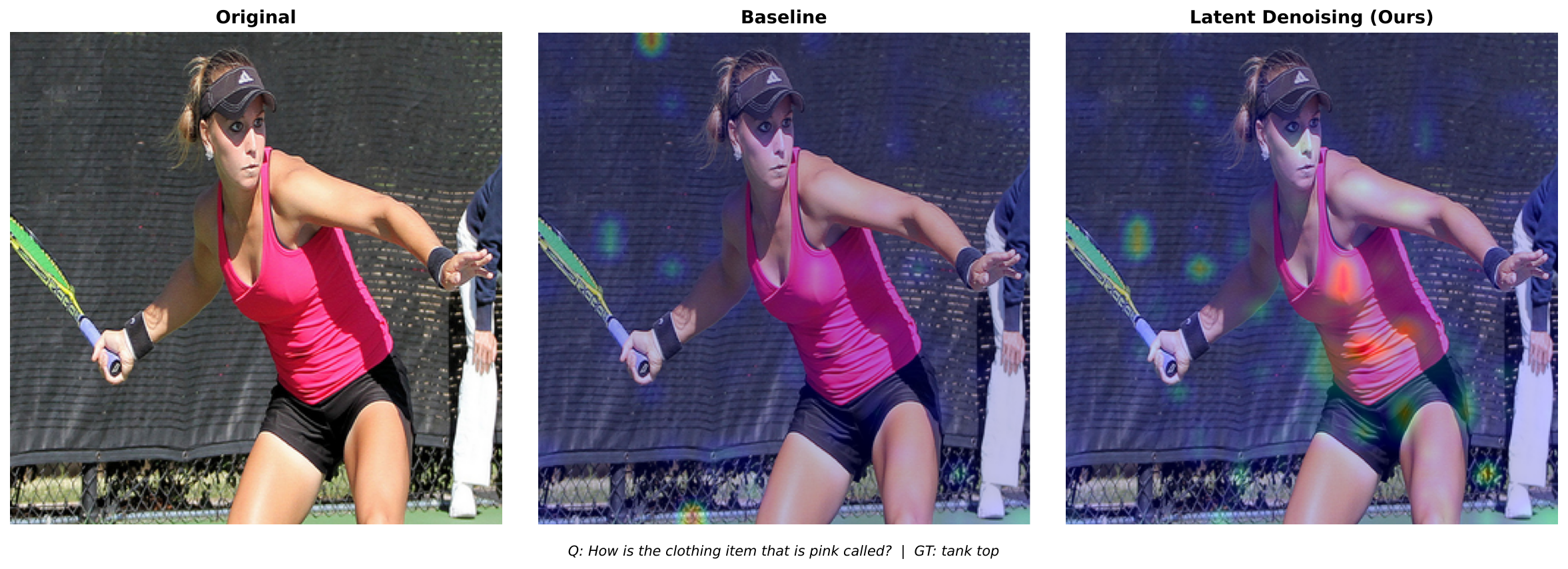}\\[-2pt]
    {\scriptsize\itshape Q: How is the clothing item that is pink called? $\mid$ GT: tank top}\\[1pt]
    \includegraphics[width=0.85\textwidth,trim=0 18 0 18,clip]{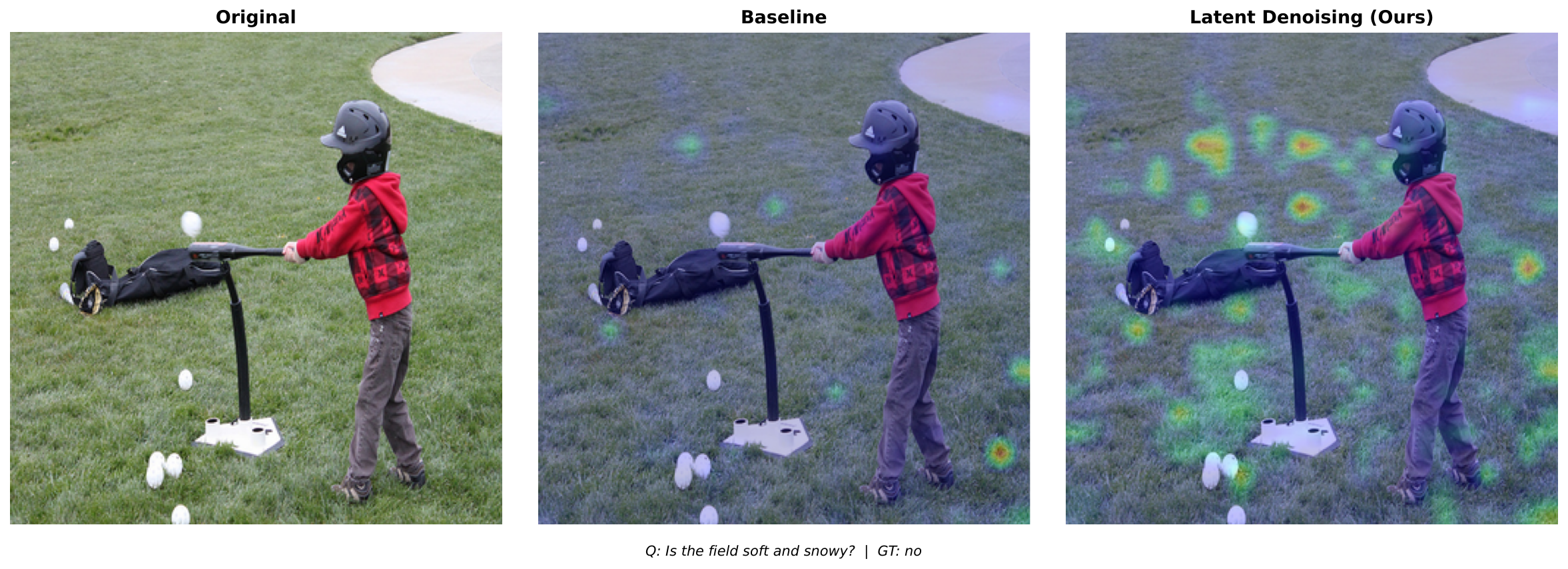}\\[-2pt]
    {\scriptsize\itshape Q: Is the field soft and snowy? $\mid$ GT: no}\\[1pt]
    \includegraphics[width=0.85\textwidth,trim=0 18 0 18,clip]{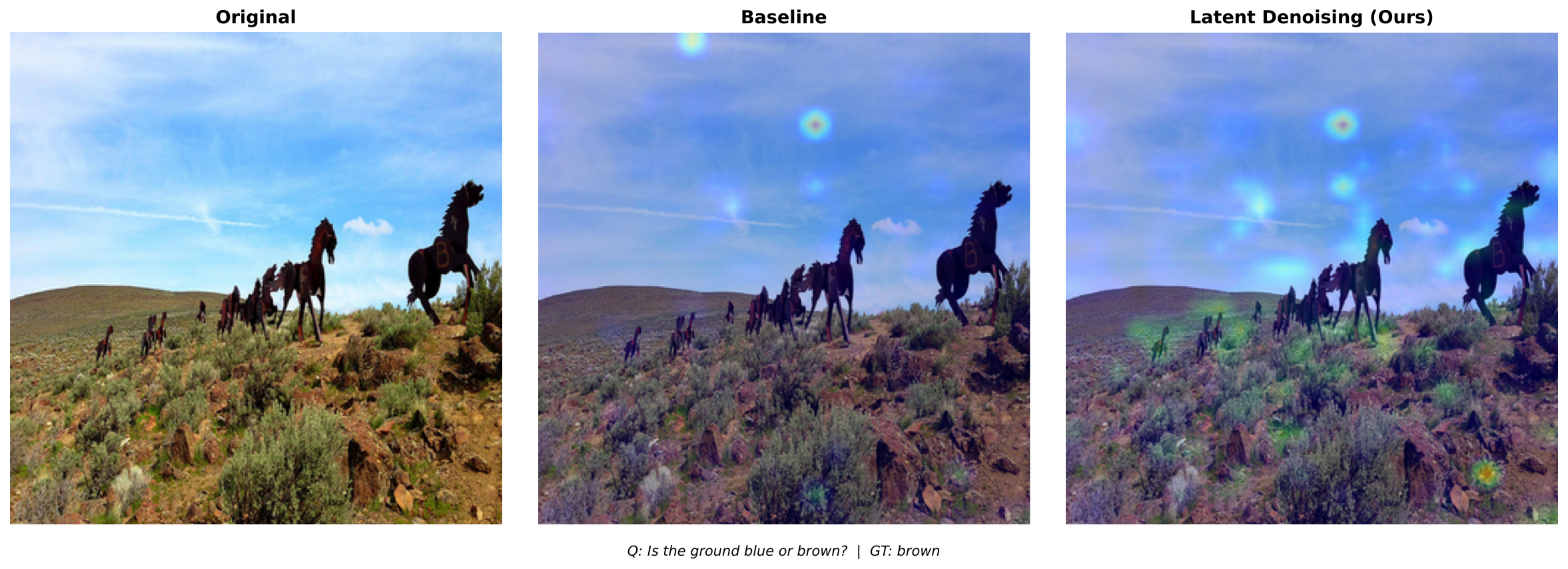}\\[-2pt]
    {\scriptsize\itshape Q: Is the ground blue or brown? $\mid$ GT: brown}
    \caption{\textbf{Additional cross-attention heatmaps (3/3).} Layer~16, GQA.}
    \label{fig:app_attn_3}
\end{figure*}

\noindent \textbf{Additional cross-attention heatmaps.}
Figures~\ref{fig:app_attn_1}--\ref{fig:app_attn_3} show additional cross-attention visualizations from layer~16 on GQA examples, where our method consistently produces more spatially grounded attention over question-relevant regions.

\clearpage

\end{document}